%% file: main.tex
\documentclass[10pt,twocolumn,letterpaper]{article}

\usepackage[pagenumbers]{cvpr} 

\input{preamble}
\definecolor{cvprblue}{rgb}{0.21,0.49,0.74}
\usepackage[pagebackref,breaklinks,colorlinks,allcolors=cvprblue]{hyperref}
\usepackage{multirow}
\usepackage[accsupp]{axessibility}
\usepackage[hang,flushmargin]{footmisc}
\input{math_commands.tex}

\def\paperID{5} 
\def\confName{CVPR}
\def\confYear{2026}

\title{A Sanity Check on Composed Image Retrieval}

\author{Yikun Liu$^{1,2}$, Jiangchao Yao$^{2\dagger}$, Weidi Xie$^{1}$, Yanfeng Wang$^{1\dagger}$ \\[3pt]
$^{1}$School of Artificial Intelligence, Shanghai Jiao Tong University, China \hspace{0.5cm} \\
$^{2}$CMIC, Shanghai Jiao Tong University, China \hspace{0.5cm}
}

\begin{document}
\twocolumn[{%
\renewcommand\twocolumn[1][]{#1}%
\maketitle

\begin{center}
   \centering
   \vspace{-15pt}
   \includegraphics[width=\textwidth]{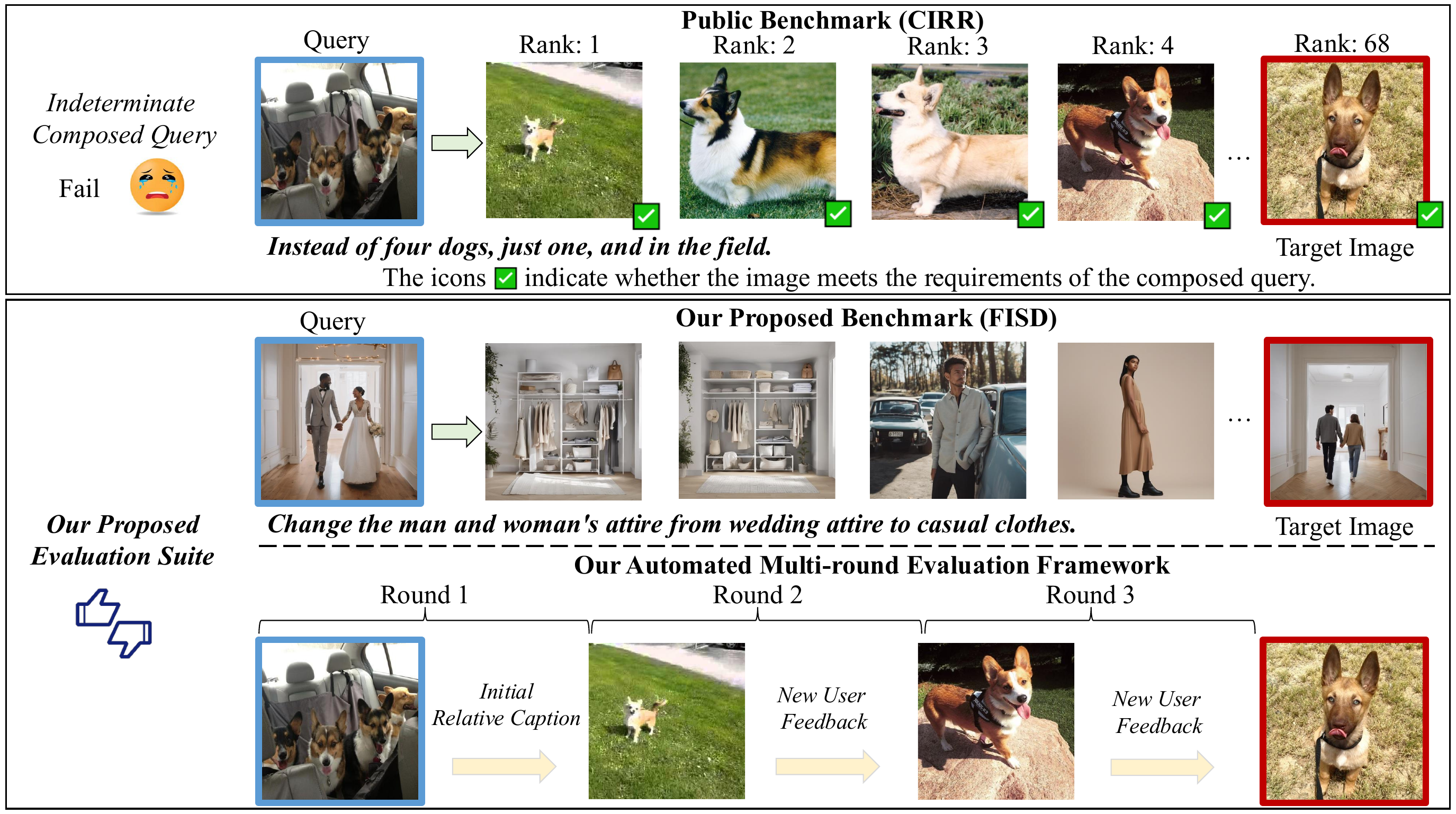}
   \vspace{-8pt}
   \captionof{figure}{\textbf{Our Motivation.} We find that existing CIR models frequently struggle to retrieve the target image when confronted with an indeterminate composed query on mainstream benchmarks. To more accurately evaluate the performance of CIR models, we propose an evaluation suite to better monitor the progress, including a novel CIR benchmark and an automated multi-round evaluation framework.
   }
  \label{fig:teaser}
 \end{center}
 }]
 
\renewcommand{\thefootnote}{\fnsymbol{footnote}} 
\footnotetext[2]{Corresponding author.} 
\renewcommand{\thefootnote}{\arabic{footnote}}
 
\input{sec/0_abstract}    
\input{sec/1_intro}
\input{sec/2_related_work}
\input{sec/3_methods}

\input{sec/4_experiments}
\input{sec/5_conclusion}

{
    \small
    \bibliographystyle{ieeenat_fullname}
    \bibliography{main}
}

\input{sec/X_suppl}

\end{document}

%% file: math_commands.tex
\usepackage{amsmath,amsfonts,bm}

\def\eqref#1{equation~\ref{#1}}

\def\1{\bm{1}}

\def\vq{{\bm{q}}}

\def\vt{{\bm{t}}}

\def\mH{{\bm{H}}}

\DeclareMathAlphabet{\mathsfit}{\encodingdefault}{\sfdefault}{m}{sl}
\SetMathAlphabet{\mathsfit}{bold}{\encodingdefault}{\sfdefault}{bx}{n}
\newcommand{\tens}[1]{\bm{\mathsfit{#1}}}

\def\tI{{\tens{I}}}

\def\sV{{\mathbb{V}}}

\newcommand{\R}{\mathbb{R}}



%% file: sec/0_abstract.tex
\begin{abstract}
Composed Image Retrieval (CIR) aims to retrieve a target image based on a query composed of a reference image, and a relative caption that specifies the desired modification. 
Despite the rapid development of CIR models, their performance is not well characterized by existing benchmarks, which inherently contain indeterminate queries degrading the evaluation (\textit{i.e.,} multiple candidate images, rather than solely the target image, meet the query criteria), and have not considered their effectiveness in the context of the multi-round system. Motivated by this, we consider improving the evaluation procedure from two aspects:
1) we introduce \textbf{FISD}, a \textbf{F}ully-\textbf{I}nformed \textbf{S}emantically-\textbf{D}iverse benchmark, which employs generative models to precisely control the variables of reference-target image pairs, enabling a more accurate evaluation of CIR methods across six dimensions, without query ambiguity; 
2) we propose an automatic multi-round agentic evaluation framework to probe the potential of the existing models in the interactive scenarios. 
By observing how models adapt and refine their choices over successive rounds of queries, this framework provides a more realistic appraisal of their efficacy in practical applications. Extensive experiments and comparisons prove the value of our novel evaluation on typical CIR methods. The project page is available \href{https://code-kunkun.github.io/SanityCIR/}{here}.
\end{abstract}

%% file: sec/1_intro.tex
\section{Introduction}
\label{sec:intro}

\noindent Recently, Composed Image Retrieval (CIR) has gained significant momentum, driven by the impressive success of Vision-Language Pre-training (VLP) methods~\cite{radford2021learning,li2022blip, li2023blip}. This task entails retrieving a target image from a query composed of a reference image, and a relative caption specifying the reference-to-target modification. In comparison to traditional image-to-image or text-to-image retrieval, CIR captures a broader and more nuanced semantic understanding of user's intentions, which is particularly advantageous in fields such as e-commerce~\cite{wu2021fashion} and internet search~\cite{zhang2024magiclens}.

Existing explorations in CIR can be broadly divided into two categories. The first set~\cite{bai2023sentence, liu2023zero, gu2023compodiff, baldrati2023composed, jang2024visual, sun2024leveraging} involves training on the standardized triplets, 
each comprising a reference image, a relative caption, and a target image. These triplets may be manually annotated with extensive human effort or automatically generated via generative models with additional human refinement. The second line~\cite{saito2023pic2word, gu2023language, baldrati2023zeroshot,suo2024knowledge} utilizes more readily accessible data, such as the image-text pairs, to learn a mapping network that projects the image feature to the text space, ultimately achieving feature fusion within the text domain. 

Despite the progress made on mainstream benchmarks like FashionIQ~\cite{wu2021fashion} and CIRR~\cite{liu2021image}, we argue that existing benchmarks fail to adequately evaluate CIR methods due to the presence of spurious samples. As shown in Figure~\ref{fig:teaser}, many mainstream benchmarks include \emph{indeterminate composed queries}\footnote{This term will be used throughout the paper to refer to the issues in existing benchmarks.}, where multiple candidates satisfy the query requirements. 
Such ambiguity undermines the accurate assessment of CIR models. 
Besides, current evaluation often focuses on the performance of models in handling one-time queries, overlooking their effectiveness in the interactive scenarios, which is practical in the multi-round system. To take steps forward, we aim to build a more accurate and comprehensive evaluation suite to better characterize the performance of the CIR models. 

In specific, to address the issue of in-determinate queries, we introduce \textbf{FISD}, a \textbf{F}ully-\textbf{I}nformed \textbf{S}emantically \textbf{D}iverse benchmark, which employs the diffusion models to generate controllable reference-target image pairs, thereby reducing the occurrence of indeterminate queries. Our FISD evaluates the CIR models across six dimensions~(cardinality, addition, negation, change, background, and complex instructions), revealing that existing CIR models still exhibit poor ability in handling negation and cardinality logic, and require more improvement.
Furthermore, to probe the potential of CIR models in interactive scenarios,
we propose an automated multi-round evaluation framework consisting of three key components: 
an off-the-shelf CIR model, a ranker, and a user simulator. Initially, the CIR model combines the reference/candidate image with the relative caption to generate a composed feature for the ranker. The ranker then iteratively updates the candidate image set based on historical interactions and the image database. The user simulator modeled by open-source vision-language foundation models evaluates whether the target image is within the top-$k$ candidates to promote search; The evaluation demonstrates that existing models, when subjected to multiple rounds of iteration, can achieve significantly improved performance, greatly surpassing their single-round performance.

In summary, our contributions are as follows: 
1) we introduce \textbf{FISD}, a novel Fully-Informed Semantically Diverse benchmark to address the ambiguity issue across six dimensions in the evaluation of existing CIR models; 
2) we propose an automated, multi-round evaluation framework, to measure the performance of CIR models under multi-round interaction. This framework employs the vision-language foundation models to simulate the user interactions, thereby generating relative caption feedback that more effectively mirrors real-world situations; 
3) our comprehensive evaluation uncovers that current CIR models have significant potential for performance improvement, particularly in terms of semantics related to cardinality and negation. 
Furthermore, CIR models exhibit significant improvements when engaged in multi-round interactions, substantially outperforming their single-round counterparts.

%% file: sec/2_related_work.tex
\section{Related Work}

\noindent \textbf{Composed Image Retrieval.} 
Composed Image Retrieval (CIR) considers retrieving a target image based on the reference image and a relative caption that describes the modification. Current methods can be primarily divided into two categories. 
The first set involves methods like CLIP4CIR~\cite{baldrati2023composed}, BLIP4CIR~\cite{liu2024bi} and SPRC~\cite{bai2023sentence} that are trained on human-annotated data or approaches~\cite{liu2023zero, gu2023compodiff, levy2024data} that use generative models to synthesize the data. These methods require the standardized triplets comprising a reference image, a relative caption, 
and a target image, using contrastive learning to minimize the distance between the composed features of the reference image and the relative caption, 
and the features of the target image. The second line, known as zero-shot CIR methods, does not rely on human-annotated triplets. 
Some of them~\cite{saito2023pic2word, gu2023language, tang2024context, baldrati2023zeroshot} train a mapping network using data in various modalities to project images into the textual feature space. By concatenating these features with those extracted from the relative captions, they can leverage the text-to-image retrieval capability of multimodal models like CLIP~\cite{radford2021learning} to retrieve target images effectively. Additionally, several efforts~\cite{yang2024ldre, tang2024reason, li2024imagine} have explored training-free methods to address this task.

\begin{figure*}[t]
    \centering
    \includegraphics[width=\textwidth]{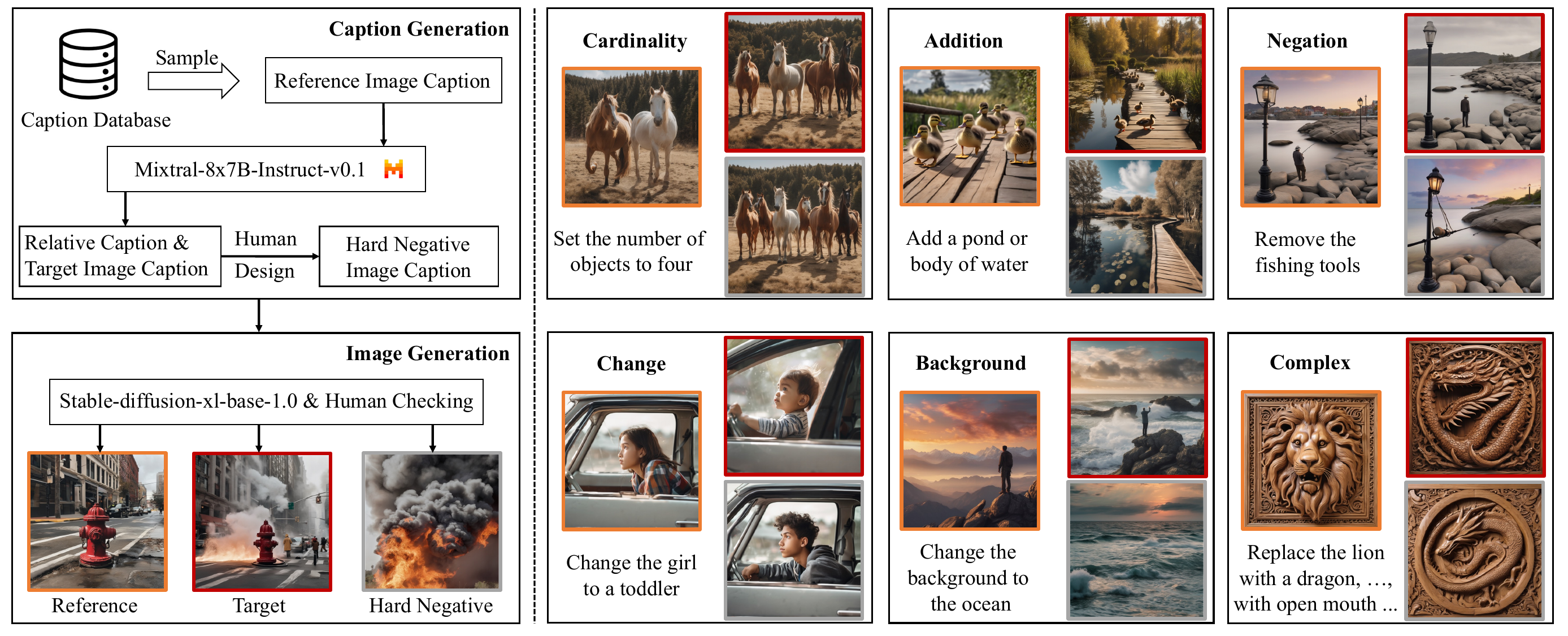}
    \caption{An overview of our FISD benchmark. The left side shows the data construction process, which includes two stages: caption generation and image generation. The right side displays examples of data for different semantic aspects, where the reference image is marked with an \textcolor{orange}{orange} frame, the target image with a \textcolor{red}{red} frame, and hard negative image with a \textcolor{gray}{gray} frame.}
    \label{fig:benchmark}
\end{figure*}

\vspace{2pt}
\noindent \textbf{Interactive Retrieval.}
The interactive image/video retrieval system is designed to iteratively capture feedback to facilitate retrieval. For example, some studies~\cite{maeoki2020interactive, madasu2022learning, liang2023simple, levy2024chatting} focus on enhancing text-to-image/video retrieval by iteratively acquiring more information in the form of visual question answering. It entails utilizing a question generator that formulates questions based on candidates from each round and historical data, followed by a user simulator to respond to these questions. Other approaches advocate providing differential feedback to describe the semantic gap between target and candidate images to improve image retrieval performance. Specifically, WhittleSearch~\cite{kovashka2017attributes} adopts a tag-based method, primarily offering feedback through attributes, whereas FashionIQ~\cite{wu2021fashion} utilizes a sentence-level method, training a relative captioner capable of generating the relative caption between candidate images and target images within the fashion domain. Inspired by these interactive methods, we propose an automated multi-round evaluation framework designed to assess how CIR models adapt and refine their selections over successive rounds of queries.

\noindent \textbf{Generative Models.} The recent significant advancements in Multimodal Large Language Models (MLLMs) and Diffusion Models have catalyzed the emergence of numerous innovative applications~\cite{wu2025mrgen, wu2025megafusion, meng2025scenegen, liu2026versavit, zhang2026transtext}. For instance, LLaVA~\cite{liu2023improved} applies MLLMs to visual question answering tasks, while LISA~\cite{lai2024lisa} utilizes MLLMs for segmentation tasks. Likewise, LamRA~\cite{liu2024lamra} leverages MLLMs for universal retrieval tasks, and LLM-as-a-judge~\cite {zheng2023judging} uses LLMs as judges to evaluate the performance of various models on open-ended questions. In parallel, StoryGen~\cite{liu2024intelligent} employs Diffusion Models to enhance visual storytelling tasks, and InstaGen~\cite{feng2024instagen} utilizes these models to synthesize data for object detection tasks. In this paper, we aim to build on this success by developing a comprehensive evaluation suite of CIR models using MLLMs and Diffusion Models. This suite aims to provide a robust framework for assessing and advancing CIR models, fostering further innovation and application in the field.

%% file: sec/3_methods.tex
\section{A New Sanity Check for CIR}

In the Composed Image Retrieval (CIR) task, 
each sample can be denoted as a triplet, 
{\em i.e.}, $\mathcal{D} = \{(\tI^{\mathrm{ref}}, \tI^{\mathrm{tgt}}, \vt) \mid \tI^{\mathrm{ref}}, \tI^{\mathrm{tgt}} \in \R^{H \times W \times 3}, \vt \in \R^{L \times 1}\}$. 
Here, $H$ and $W$ denote the height and width of an image,
and $L$ refers to the sequence length of the relative caption. Specifically, the model takes the reference image~($\tI^{\mathrm{ref}}$) and relative caption~($\vt$) as input, and construct a composed query~$\vq$, which is further used to retrieve the target image~($\tI^{\mathrm{tgt}}$) from the set~($\Omega = \{\tI_i, i = 1, \cdots , N\}$), where $N$ represents the image number in the database. 

\input{tables/fisd_eval_final}

In the literature, significant advancements~\cite{liu2024lamra, bai2023sentence, baldrati2023composed} have been made in common benchmarks, for example, CIRR~\cite{liu2021image} and FashionIQ~\cite{wu2021fashion}.
However, as illustrated in Figure~\ref{fig:teaser} and Section F of the supplementary material, we argue that the indeterminate composed queries, which correspond to numerous candidate images in popular CIR datasets, 
actually pose challenges for precise evaluation. 
To comprehensively characterize the performance of current CIR models, we explore building a more comprehensive evaluation suite to better measure CIR models, 
encompassing a novel \textbf{CIR benchmark} and an automated \textbf{multi-round evaluation framework}.

\subsection{Fully-Informed Semantically-Diverse Benchmark}\label{sc_benchmark}

For our benchmark, we consider leveraging Large Language Models (LLMs) and Diffusion Models (DMs) to synthesize the samples, since it enables us to flexibly control the variable between paired reference-target images, allowing for an exact evaluation without the spurious samples. In addition, we focus on six categories (cardinality, addition, negation, change, background, and complex instruction) to cover different real-world retrieval types, and perform careful human verification to guarantee quality. In the following, we will present the FISD construction process, which consists of caption generation and image generation as illustrated in Figure~\ref{fig:benchmark}.

\vspace{2pt}
\noindent \textbf{Caption Generation.} To obtain the triplet sample for CIR, on the textual side, we need the reference image caption, relative caption, and target image caption. For the reference image captions, we select from existing image caption datasets, such as COCO~\cite{lin2014microsoft}, CC3M~\cite{sharma2018conceptual}, Flickr30k~\cite{young2014image}, etc. Then, with this caption for the reference image, we prompt Mixtral-8x7B-Instruct-v0.1~\cite{jiang2024mixtral} to generate the relative caption and the target image caption, with a designed prompt. 
For more details about the prompts, we refer the reader to Section A of the supplementary material. 

\vspace{2pt}
\noindent \textbf{Image Generation.} After obtaining captions, we leverage Stable-Diffusion-xl-base-1.0~\cite{podell2023sdxl} to generate the images. To ensure consistency between the reference and target images during this process. 
To minimize the risk of false negatives, we employ the same random seed. Furthermore, to narrow the gap between the generated and natural images, we \emph{manually} curate the dataset by excluding cartoon-style images, logically flawed images, or those that are overtly unrealistic. 
Further analysis is provided in Section~\ref{sec:ablate}. In total, we obtain 200 triplets for each category, resulting in a comprehensive dataset of 1200 triplets~(3600 images).

To challenge the existing CIR models, and avoid trivial solutions,
such as retrieving the target image based solely on image similarity, we generate hard negative images for each triplet. These images are crafted by \emph{manually} adjusting the corresponding caption to ensure it closely resembles either the target image or the relative caption.
For instance, in the ``change'' category, the relative caption can be \textit{``replacing the reference image with a girl wearing long pants''}, 
while the hard negative images might be \textit{``\textit{a girl wearing shorts or a boy wearing long pants}''}. This characterizes the challenge of our FISD benchmark, where the model must fully comprehend both the reference image and the relative caption to effectively avoid retrieving the hard negative images.

\subsection{Multi-round Evaluation Framework}

Conventional evaluation metric often emphasizes the performance of models in managing single queries, neglecting the dynamic and evolving context that characterizes real-world applications. To overcome this limitation, we propose that the evaluation of CIR models should incorporate multi-turn interactions. Ideally, multi-turn evaluations would involve human participation. If a query in the current round does not retrieve the target image, the user should be able to provide further feedback based on the retrieved images, and this process would continue iteratively.

However, employing humans for multi-round evaluation can be challenging and inconvenient. Therefore, we propose an \textbf{automated multi-round evaluation framework} to understand how the CIR models adapt and refine their choices over successive rounds of queries. 
Our evaluation framework is composed of three main components: 
a generic off-the-shelf CIR model, a ranker, and a user simulator intended to mimic user feedback. In the subsequent sections, we present a detailed description of the workflow and the critical components of this framework. 

\begin{figure*}[ht]
    \centering
    \includegraphics[width=\textwidth]{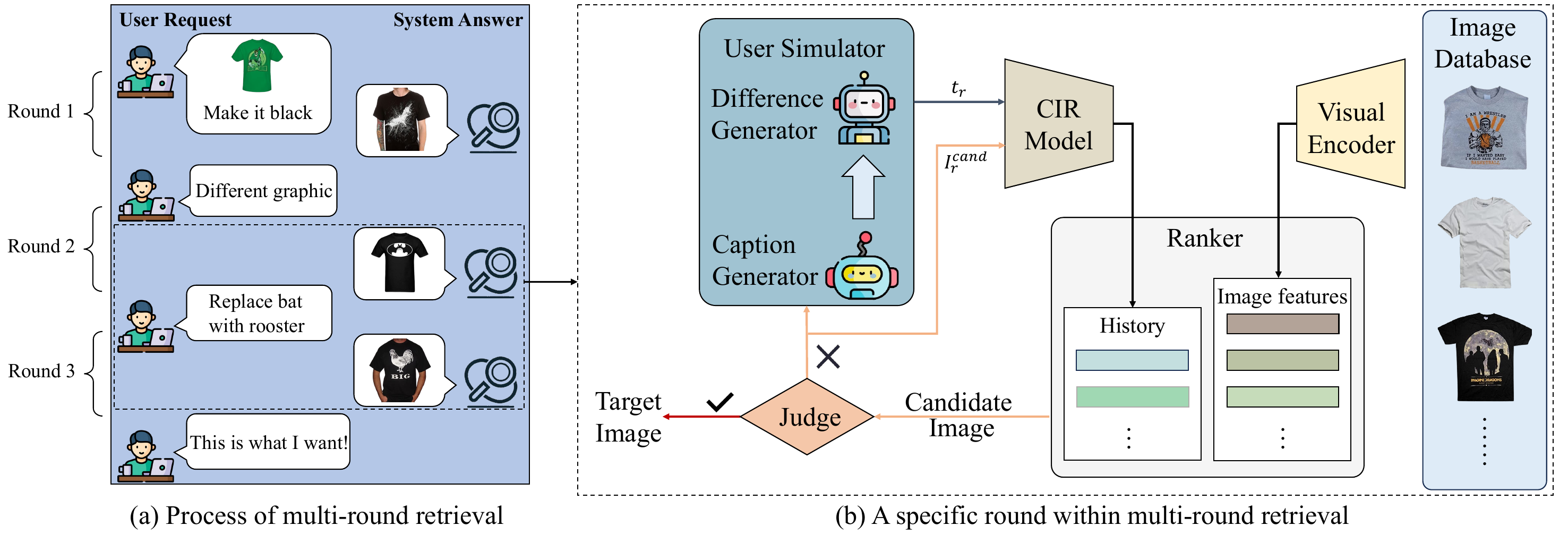}
    \caption{\textbf{Overview of our automated multi-round evaluation framework.} The user initially provides a reference image and the relative caption. Subsequently, the CIR model takes these inputs to generate a composed query feature, which is then stored in the history list. Next, the ranker uses the history list and all image features to determine candidate images. Finally, the selected candidate image becomes the reference image for the next round and is fed into the user simulator to generate the relative caption for the subsequent round.}
    \label{fig:architecture}
    \vspace{-8pt}
\end{figure*}

\vspace{2pt}
\noindent \textbf{Interaction Workflow.} 
As shown in Figure~\ref{fig:architecture}, a user first inputs a reference image and a relative caption into the retrieval system for searching a target image. Then, the system outputs candidate images, which contain $M$ items. This $M$ can be varied. If the target image is included in the top-$k$ candidate images, the retrieval is successfully finished. If not, the user simulator receives the candidate images and the target image to output a relative caption, which describes the difference between the candidate image and the target image. 
Here, this candidate image can vary from top-1 to top-$M$, but we specifically use the top-1 candidate. Such a loop is iteratively performed until the user reaches the target image in the top-$k$ candidate images or the round of interaction reaches a predetermined upper limit $R_{\mathrm{max}}$.

\vspace{2pt}
\noindent \textbf{CIR Model.} Considering the \(r\)-th round interaction~$(r = 1, \cdots, R_{\mathrm{max}})$, the CIR model takes the candidate image~$\tI_{r}^{\mathrm{cand}}$ and the relative caption~$\vt_r$ as inputs, and construct a composed query feature~$\vq_r = \texttt{fusion}(f_{\mathrm{visual}}(\tI_r^{\mathrm{cand}}), f_{\mathrm{text}}(\vt_{r}))\in \mathbb{R}^{1 \times d}$, where $\texttt{fusion}(\cdot, \cdot)$ refers to fusion method employed within CIR model, $f_{\mathrm{visual}}$ denotes the visual encoder for extracting image features, $f_{\mathrm{text}}$ represents the text encoder for extracting text embeddings, and $d$ refers to the feature dimension. Following this, we utilize the visual encoder to encode all images~$\{\tI_i\}_{i=1}^{N}$ in the image database to get the image features~$\sV = \{v_i\}_{i=1}^{N} $, 
where $v_i \in \R^{1 \times d}$. Finally, both the composed query feature~$\vq_r$ and image features~$\sV$ are fed into the ranker to perform the ranking. It is worth noting that any off-the-shelf CIR model can be adapted to our framework.

\vspace{2pt}
\noindent \textbf{Ranker.} The ranker receives $r$-th round's composed query feature~$\vq_r$ along with all image features~$\sV$. Furthermore, the ranker maintains a history list~$\mH = [\vq_1, \cdots, \vq_r]$, which aggregates the composed query feature from each round. This facilitates the tracking of query features across different rounds. Subsequently, the ranker computes the distance between the history representation \( f_h(\mH) \) and each image feature vector \( v \in \sV \) using the equation:
\begin{equation}
    \tI_{r+1}^{\mathrm{cand}} = \underset{v \in \sV}{\mathrm{argmin}} \, \texttt{sim}(f_h(\mH), v). 
\end{equation}
Here, \( f_h \) represents a fusion function applied to the history list, specifically a simple averaging operation defined by \( f_h(\mH) = \frac{1}{r} \sum_{i=1}^r \vq_i \), where the average is computed over all elements from the first round up to the current $r$-th round. \( \texttt{sim}(\cdot, \cdot) \) denotes the cosine distance. Finally, the ranker utilizes a greedy selection strategy, opting for the image with the smallest cosine distance as the candidate.

\input{tables/multi_round_eval}

\vspace{2pt}
\noindent \textbf{User Simulator.} The user simulator operates as an oracle, adept at identifying differences between the candidate image $\tI_{r+1}^{\mathrm{cand}}$ and the target image $\tI^{\mathrm{tgt}}$~\cite{wu2021fashion}. Here, to mimic human interactions, we consider integrating the pre-trained foundation models, including the Multimodal Large Language Model (MLLM) and Large Language Model (LLM). Concretely, with MLLM and LLM, the user feedback~$\vt_{r+1}$ is provided in the form of a relative caption, which can be formally represented as follows:
\begin{equation}
\vt_{r+1} = {\texttt{LLM}} \left( \texttt{MLLM}(\tI_{r+1}^{\mathrm{cand}}), {\texttt{MLLM}}(\tI^{\mathrm{tgt}}) \right).
\end{equation}
The process involves generating detailed captions for both images using the ${\texttt{MLLM}}$. These captions are then processed by the ${\texttt{LLM}}$ to generate user feedback that highlights the discrepancy. Upon generating user feedback, it is combined with the candidate image as input into the CIR model for subsequent iterations. This process is iterated until either of two conditions is met: the target image is ranked within the top-$k$ results, namely, a successful retrieval, or the maximum number $R_{max}$ is reached. For detailed prompts, please refer to Section B of the supplementary material.

%% file: tables/fisd_eval_final.tex
\begin{table*}[!t]
\scriptsize
\setlength{\tabcolsep}{0.3cm}
\centering
\caption{Performance of various CIR models on FISD benchmark.}
\resizebox{\textwidth}{!}{
\begin{tabular}{l c c c c c c c}
\toprule
 \multicolumn{1}{c}{~}&
 \multicolumn{1}{c}{\bf Cardinality}& 
 \multicolumn{1}{c}{\bf Addition}& 
 \multicolumn{1}{c}{\bf Negation}&
 \multicolumn{1}{c}{\bf Change}&
 \multicolumn{1}{c}{\bf Background}&
 \multicolumn{1}{c}{\bf Complex Inst.}&
 \multicolumn{1}{c}{\bf Average}\\
 \cmidrule(r){2-2} \cmidrule(r){3-3} \cmidrule(r){4-4} \cmidrule(r){5-5} \cmidrule(r){6-6} \cmidrule(r){7-7} \cmidrule(r){8-8}
 Model & \text{Recall@1} & \text{Recall@1} & \text{Recall@1} & \text{Recall@1} & \text{Recall@1} & \text{Recall@1} & \text{Recall@1}\\ \midrule
 Pic2Word~\cite{saito2023pic2word} & 6.00 & 39.00 & 0.00 & 24.00 & 18.00 & 41.00 & 21.33 \\ 
 Context-I2W~\cite{tang2024context} & 14.00  & 48.00 & 0.00 & 29.50 & 27.50 & 48.00 & 27.83 \\
 LinCIR~\cite{gu2023language} & 5.50 & 52.50 & 1.50 & 43.50 & 36.50 & 55.00 & 32.42 \\
 TransAgg~\cite{liu2023zero} & 15.50 & 56.00 & 2.00 & 52.00 & 26.00 & 57.00 & 34.75 \\
CLIP4CIR~\cite{baldrati2023composed} & 31.00 & 67.50 & 4.00 & 65.00 & 43.50 & 72.00 & 47.17 \\ 
 BLIP4CIR+Bi~\cite{liu2024bi} & 35.50 & 65.50 & 3.50 & 58.50 & 42.50 & 65.50 & 45.17 \\
TG-CIR~\cite{wen2023target} & 43.50 & 69.00 & 1.50 & 64.50 & 47.50 & 65.50 & 48.58 \\
CoVR$^{*}$~\cite{ventura2024covr} & 47.50 & 70.00 & 8.50 & 77.00 & 61.50 & 70.00 & 55.75 \\
CIReVL~\cite{karthik2023vision} & 24.00 & 53.50 & 17.50 & 61.00 & 40.00 & 69.00 & 44.17 \\
 SPRC~\cite{bai2023sentence} & 41.00 & 66.50 & 9.50 & 73.00 & 48.50 & 62.00 & 50.08 \\ 
 SPN4CIR~\cite{feng2024improving} & 58.50 & 69.50 & 13.00 & 77.00 & 56.00 & 61.00 & 55.83 \\ 
 \bottomrule
 \end{tabular}}
\label{table:benchmark}
\end{table*}

%% file: tables/multi_round_eval.tex
\begin{table*}[t]
\centering
\scriptsize 
\caption{Multi-round evaluation on various state-of-the-art CIR models across a range of benchmarks.}
\resizebox{\textwidth}{!}{
\setlength{\tabcolsep}{0.26cm}
\begin{tabular}{l c c c c c c c c c c}
\toprule
 \multirow{2}{*}[-0.5ex]{Method}&
 \multicolumn{2}{c}{\textbf{FashionIQ-Dress}}& 
 \multicolumn{2}{c}{\textbf{FashionIQ-Shirt}}&
 \multicolumn{2}{c}{\textbf{FashionIQ-Toptee}}&
 \multicolumn{2}{c}{\textbf{CIRR}}&
 \multicolumn{1}{c}{\textbf{CIRCO}}&
 \multicolumn{1}{c}{\textbf{FISD}}\\
 \cmidrule(lr){2-3} \cmidrule(lr){4-5} \cmidrule(lr){6-7} \cmidrule(lr){8-9} \cmidrule(lr){10-10} \cmidrule(lr){11-11}
  & Hits@10 & Hits@50 & Hits@10 & Hits@50 & Hits@10 & Hits@50 & Hits@1 & Hits@5 & MAP@5 & Hits@1\\
 \midrule
\multicolumn{11}{c}{\textit{Round 1}} \\
\midrule 
  Pic2Word~\cite{saito2023pic2word} & 20.00 & 40.20 & 26.20 & 43.60 & 27.90 & 47.40 & 23.25 & 51.42 & 7.39 & 21.33 \\
  Context-I2W~\cite{tang2024context} & 23.10 & 45.30 & 29.70 & 48.60 & 30.60 & 52.90 & 26.96 & 56.59 & 11.96 & 27.83\\
  LinCIR~\cite{gu2023language} & 20.92 & 42.44 & 29.10 & 46.81 & 28.81 & 50.18 & 25.09 & 54.41 & 10.61 & 32.42\\
  TransAgg~\cite{liu2023zero} & 30.24 & 51.91 & 34.45 & 53.97 & 38.40 & 59.51 & 38.79 & 69.58 & 11.06 & 34.75\\
  CLIP4CIR~\cite{baldrati2023composed} & 39.46 & 64.55 & 44.41 & 65.26 & 47.48 & 70.98 & 45.37 & 78.47 & 10.35 & 47.17 \\ 
BLIP4CIR+Bi~\cite{liu2024bi} & 42.09 & 67.33 & 41.76 & 64.28 & 46.61 & 70.32 & 42.36 & 75.48 & 10.09 & 45.17  \\
 SPRC~\cite{bai2023sentence} & 49.18 & 72.43 & 55.64 & 73.89 & 59.35 & 78.58 & 55.39 & 84.26 & 21.17 & 50.08  \\ 
 SPN4CIR~\cite{feng2024improving} & 50.57 & 74.12 & 57.70 & 75.27 & 60.84 & 79.96 & 56.47 & 85.29 & 21.78 & 55.83 \\ 
   \midrule
\multicolumn{11}{c}{\textit{Round 3}} \\
\midrule
 Pic2Word~\cite{saito2023pic2word} & 29.65 & 55.08 & 39.99 & 59.76 & 39.16 & 61.75 & 40.28 & 71.42 & 12.93 & 64.67\\
  Context-I2W~\cite{tang2024context} & 39.86 & 64.20 & 49.71 & 69.28 & 51.35 & 73.79 & 43.05 & 75.27 & 21.14 & 50.25 \\
  LinCIR~\cite{gu2023language} & 33.66 & 59.94 & 51.03 & 69.92 & 48.24 & 68.59 & 43.03 & 75.68 & 17.64 & 51.08\\
  TransAgg~\cite{liu2023zero} & 48.79 & 72.88 & 60.06 & 79.34 & 65.73 & 83.53 & 61.83 & 89.38 & 27.22 & 77.67\\
  CLIP4CIR~\cite{baldrati2023composed} & 53.74 & 77.99 & 65.95 & 85.18 & 68.08 & 87.10 & 67.54 & 95.72 & 23.57 & 74.25\\
BLIP4CIR+Bi~\cite{liu2024bi} & 59.54 & 81.95 & 57.56 & 79.49 & 64.41 & 83.02 & 54.99 & 89.48 & 17.70 & 70.58\\
 SPRC~\cite{bai2023sentence} & 67.38 & 85.52 & 77.13 & 90.43 & 78.74 & 91.74 & 81.37 & 97.32 & 36.86 & 77.08\\
 SPN4CIR~\cite{feng2024improving} & 70.70 & 88.00 & 80.37 & 92.35 & 82.51 & 93.01 & 84.31 & 97.66 & 38.02 & 81.00\\
\midrule
\multicolumn{11}{c}{\textit{Round 5}}\\
\midrule
Pic2Word~\cite{saito2023pic2word} & 33.12 & 59.05 & 43.42 & 64.13 & 42.22 & 65.37 & 57.47 & 75.60 & 17.19 & 83.50 \\
  Context-I2W~\cite{tang2024context} & 44.42 & 68.12 & 55.69 & 74.19 & 56.81 & 78.68 & 56.83 & 78.71 & 26.89 & 69.08\\
  LinCIR~\cite{gu2023language} & 37.04 & 62.96 & 56.43 & 74.93 & 51.91 & 72.41 & 58.48 & 79.14 & 21.51 & 68.92\\
  TransAgg~\cite{liu2023zero} & 53.59 & 76.85 & 67.12 & 84.25 & 73.33 & 89.14 & 78.04 & 92.54 & 32.26 & 88.33\\
  CLIP4CIR~\cite{baldrati2023composed} & 57.86 & 81.51 & 70.71 & 88.42 & 72.67 & 89.95 & 82.92 & 97.06 & 29.32 & 82.08\\
BLIP4CIR+Bi~\cite{liu2024bi} & 63.41 & 84.28 & 61.83 & 82.09 & 67.87 & 86.18 & 67.11 & 91.32 & 20.71 & 80.42\\
 SPRC~\cite{bai2023sentence} & 70.40 & 87.70 & 80.72 & 92.54 & 83.32 & 93.98 & 89.76 & 98.49 & 41.93 & 84.92\\
 SPN4CIR~\cite{feng2024improving} & 75.21 & 90.83 & 84.45 & 94.80 & 87.25 & 95.72 & 91.32 & 98.76 & 40.79 & 86.92\\
\bottomrule
\end{tabular}}
\label{table:multi_round_table}
\end{table*}

%% file: sec/4_experiments.tex
\section{Experiments}\label{sec_exp}

\subsection{Experimental Setups}\label{subsec_experimental_setup}

\noindent \textbf{Datasets and Evaluation Metric.}
We evaluate various CIR models on four benchmarks, including three public benchmarks: CIRR~\cite{liu2021image}, FashionIQ~\cite{wu2021fashion}, CIRCO~\cite{baldrati2023zeroshot}, and our own proposed FISD benchmark. For CIRR and CIRCO, we conduct evaluations on the validation set because the multi-round setting requires the involvement of the target image, and the ground truth for the test set is not accessible. In the context of multi-round evaluations, we primarily adopt the standard metrics in multi-round retrieval, {\em i.e.}, Hits@K, where success is defined as the target image appearing in the top-$K$ results in any round up to the current one. Moreover, we include the rank of the target image as an evaluation metric. Additional experiments involving Recall@K metrics under the multi-round setting are included in Section D of the supplementary material.

\vspace{2pt}
\noindent \textbf{Implementation Details.} The evaluation process on the FISD is meticulously designed to distinguish among various semantic subsets, each of which constitutes an independent image database comprising 600 images per subset. Our multi-round evaluation framework is implemented with PyTorch and is compatible with off-the-shelf CIR models such as Pic2Word~\cite{saito2023pic2word}, LinCIR~\cite{gu2023language}, and SPRC~\cite{bai2023sentence}. These models are evaluated using their official checkpoints. By default, we utilize llama3-llava-next-8b~\cite{liu2024llavanext} and Meta-Llama3-8B-Instruct~\cite{dubey2024llama} as MLLM and LLM respectively to construct our user simulator. The maximum number of rounds, denoted as $R_{max}$, is set to 5. Experiments detailed on Section~\ref{sec:ablate} are primarily conducted using the SPRC model. The entire process is implemented on an Nvidia A100 GPU with 80GB of memory. For more implementation details, please refer to Section C of the supplementary material.

\subsection{Main Results}

\noindent \textbf{Evaluation on FISD Benchmark.} We evaluate various CIR models across diverse backbones, training data, and methodologies using the FISD benchmark. As shown in Table \ref{table:benchmark}, we can draw the following conclusions: \textbf{(i)} \textit{Current CIR models exhibit notable deficiencies in several semantic aspects, particularly in processing negation and cardinality semantics.} For instance, regarding the semantics of negation, even the best-performing model in our evaluation achieves only a Recall@1 of 17.5. This may be attributed to the inherent limitations of the feature extraction backbone, which performs inadequately in these semantic aspects~\cite{singh2024learn, paiss2023teaching}.\textbf{(ii)} \textit{Current CIR models perform relatively well at the more ``direct'' semantic aspect.} When dealing with semantics such as ``addition'' and ``change'', the relative caption often explicitly identifies the elements present in the target image. In these scenarios, existing models perform slightly better. For example, SPN4CIR achieves a recall@1 of 77 on the change subset. \textbf{(iii)} \textit{Current CIR models have room for improvement across various semantic dimensions.} Despite the provision of precise details, the challenge of understanding certain semantics remains unresolved in current CIR methods. For example, the best-performing model we evaluated achieved an average Recall@1 of only 55.83, highlighting the need for further enhancement.

\vspace{2pt}
\noindent \textbf{Multi-round Evaluation Results.} 
To assess the performance of existing CIR models in a multi-round setting, we utilize our proposed automated multi-round evaluation framework for eight different publicly available CIR models. These models encompass a variety of training data and strategies. The experimental results are presented in Table~\ref{table:multi_round_table}. 
We can draw the following observations: 
\textbf{(i)} \textit{Multi-round interactions can significantly enhance performance.} The experimental results demonstrate that regardless of the initial performance of the CIR models, they achieve significant improvements in performance across all four benchmarks in a multi-round setting. For instance, with the Pic2Word model on the CIRR dataset, the Hits@1 in the first round is merely 23.25, but after three rounds of interaction, the result improves to 40.28, an increase of approximately 73.25\%. Similarly, for SPN4CIR on the CIRR dataset, the single-round Hits@5 is 85.29, but after five-round processing, the Hits@5 soars to 98.76. 
\textbf{(ii)} \textit{The effectiveness of a CIR model in a multi-round setting is closely linked to its initial performance.} Generally, models with strong initial performance tend to achieve better outcomes following multiple rounds of interaction. 
\textbf{(iii)} \textit{The performance gains achieved through multi-round interactions tend to diminish as the number of interactions increases.} Specifically, the improvement observed after three rounds is greater than the improvement seen when progressing from three to five rounds. For example, with the SPN4CIR on the FashionIQ-Dress, the Hits@10 improved by 20.23 points from the first to the third round, whereas it increased by only 4.51 points from the third to the fifth round.

\subsection{Further Analysis}
\label{sec:ablate}

\begin{figure}
    \centering
\includegraphics[width=.5\textwidth]{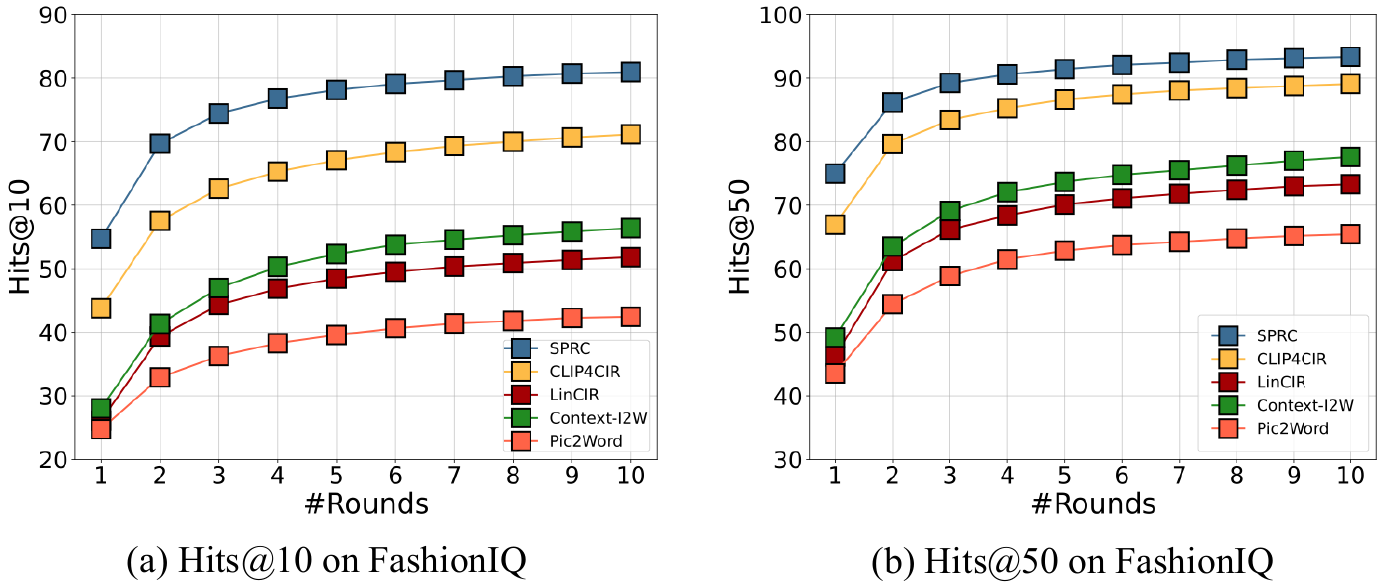}
    \caption{Performance across various rounds on FashionIQ.}
    \label{fig:multi_round}
\end{figure}

\noindent \textbf{Analysis of Multiple Rounds.} 
As shown in Figure \ref{fig:multi_round}, we investigate the changes in retrieval performance across different maximum rounds on FashionIQ. The results indicate that performance consistently improves as the number of rounds increases, demonstrating the effectiveness of current CIR models in enhancing performance over multiple iterations. Notably, the performance gains are more pronounced at lower rounds. As the number of rounds exceeds 5, the performance improvements gradually plateau. Therefore, we set the default maximum rounds $R_\mathrm{max}$ to 5.

\input{tables/ablation_mllm_llm}

\vspace{2pt}
\noindent \textbf{Impact of MLLM\&LLM on Evaluation Performance.} In this part, we examine the impact of utilizing various MLLMs and LLMs as user simulators on the performance of multi-round evaluations based on a particular CIR model, SPRC.
We utilize BLIP-2~\cite{li2023blip}, LLaVA-1.5~\cite{liu2023improved}, and LLaVA-Next~\cite{liu2024llavanext} as MLLMs, alongside Llama2-7B~\cite{touvron2023llama}, Mistral-7B~\cite{jiang2023mistral}, and Llama3-8B~\cite{dubey2024llama} as LLMs. The experimental results are shown in Table~\ref{tab: MLLM_LLM}. We can make the following observations: \textbf{(i)} \textit{Stronger MLLMs yield better outcomes.} Specifically, BLIP-2 tends to generate coarse captions, whereas captions produced by LLaVA-1.5 and LLaVA-Next are more detailed, leading to superior performance. \textbf{(ii)} \textit{Models with stronger instruction-following capabilities are more suitable for use as user simulators.} There are noticeable improvements with Mistral-7B and Llama3-8B over Llama2-7B, likely due to Llama2-7B's weaker instruction-following capabilities that often result in irrelevant content and thus mislead the model. Conversely, Mistral-7B and Llama3-8B show robust instruction-following abilities, and there is no significant difference in performance between them. \textbf{(iii)} \textit{Our evaluation framework leads to a consistent improvement using different combinations of MLLMs and LLMs.} Despite some variations in the performance caused by different combinations of MLLMs and LLMs, there is a significant improvement compared to single-round results.

\input{tables/qwen2vl}

\vspace{2pt}
\noindent \textbf{Different Simulation Method.} 
As MLLMs advance, they can now directly identify differences between images. Leveraging this functionality, we employ Qwen2-VL-7B~\cite{wang2024qwen2} to directly generate the difference between two images as our user simulation. As shown in Table~\ref{qwen2vl}, this approach also achieves significant performance improvements. Compared to using LLMs to infer differences from captions, this simulation method may be more suitable when captions are missing or of low quality, though it incurs slightly higher inference costs.

\input{tables/user_study}

\vspace{2pt}
\noindent \textbf{User Study.} To assess the multi-round performance of the CIR model in real-world application scenarios, we conduct user studies across three benchmarks. Specifically, we select 50 samples that exhibit poor performance in single-round CIR models from each benchmark. These samples undergo 5 interaction rounds with both real users and our user simulator. As illustrated in Table~\ref{tab:user_study}, it is evident that the ranking of the target images significantly decreases, regardless of whether feedback is provided by a user or the simulator. For instance, the ranking of the target image falls by approximately 2000 positions after 5 interaction rounds using either method. This underscores the versatility and practicality of our framework. It is noteworthy that, at times, user performance is inferior to that of the user simulator. 
This arises because users tend to provide short feedback, whereas simulators are prone to offer more detailed feedback, as discussed in Section E of the supplementary materials. The nature of feedback difference inherently makes some impact on the effectiveness of interaction. Overall, our user simulator closely aligns with the performance of real users, reflecting actual evaluation conditions.

\input{tables/ablation_studies}

\vspace{2pt}
\noindent \textbf{Ablation Studies of our Evaluation Framework.} In this section, we conduct ablation studies on CIRR to evaluate three key designs: \textbf{(i)} The effect of history information. \textbf{(ii)} The role of varied feedback. \textbf{(iii)} The rationale for selecting the top-1 candidate images as the next round's reference. The experimental results in Table~\ref{ablation} lead to the following conclusions: 
\textbf{(i)} Using candidate image feature at round $r$ directly (without averaging with previous rounds) degrades performance, confirming the effectiveness of historical information. \textbf{(ii)} Reusing the same relative caption across varying candidate images significantly harms performance, demonstrating the importance of diverse feedback. \textbf{(iii)} Randomly selecting a candidate image from the top-10 yields negligible performance differences, justifying the use of the top-1 candidate for the next round.

\vspace{2pt}
\noindent \textbf{Can FISD Accurately Reflect the CIR Model's Capability?} Since the images in the FISD benchmark are all synthetic, there remains a gap between these and natural images. This raises the question: can FISD fully capture the performance of CIR models? As shown in Table~\ref{table:benchmark}, CIR models perform poorly in handling negation semantics. To determine whether this issue is caused by the gap between synthetic and natural images, we use Llama3-8B to select and manually verify queries with negation semantics from the CIRR validation set. The experimental results, presented in Table 1 of the supplementary material, reveal that even with natural images, CIR models struggle with negation compared to other semantic aspects. This is consistent with our findings on FISD. Furthermore, the performance trends across different CIR models on the FISD align with those on CIRR and FashionIQ. Therefore, we believe that FISD can accurately reflect the capabilities of CIR models.

%% file: tables/ablation_mllm_llm.tex
\begin{table}
    \centering
    \scriptsize
    \caption{Performance of various MLLMs and LLMs on CIRR.}
    \setlength{\tabcolsep}{4mm}{
    \resizebox{.45\textwidth}{!}{
    \begin{tabular}{cccc}
    \toprule
    MLLM & LLM & Hits@1  & Hits@5\\
    \midrule
    BLIP-2 & Llama3-8B & 85.82 & 97.54\\
    LLaVA-1.5 & Llama3-8B  & 89.62 & 98.37\\
    LLaVA-Next & Llama3-8B  & \textbf{89.76} &  \textbf{98.49} \\
    \midrule
    LLaVA-Next & Llama2-7B & 78.43 & 94.07\\
    LLaVA-Next & Mistral-7B  & 89.45 & \textbf{98.64}\\
    LLaVA-Next & Llama3-8B  & \textbf{89.76} &  98.49 \\
    \bottomrule
    \end{tabular}}}
    \label{tab: MLLM_LLM}
    \vspace{-8pt}
\end{table}

%% file: tables/qwen2vl.tex
\begin{table}[h]
\setlength{\tabcolsep}{0.4mm}
\scriptsize 
\caption{Comparison of various simulation methods on CIRR. R1 represents Round 1.}
\resizebox{\linewidth}{!}{
\begin{tabular}{lccc}
\toprule
Method & R1/(Hits@1$\uparrow$/Rank$\downarrow$) & R3/(Hits@1$\uparrow$/Rank$\downarrow$) & R5/(Hits@1$\uparrow$/Rank$\downarrow$) \\
\midrule
Qwen2VL & 55.39 / 6.21  & 80.80 / 2.01 & 90.00 / 1.53 \\
Llama3 & 55.39 / 6.21  & 81.37 / 1.79 & 89.76 / 1.45 \\
\bottomrule
 \end{tabular}
}
\label{qwen2vl}
\end{table}
\vspace{-5pt}

%% file: tables/user_study.tex
\begin{table}
    \centering
    \scriptsize
     \caption{Performance of real users and our user simulator. Our user simulator aligns with the performance of the real users.}
    \setlength{\tabcolsep}{3mm}{
    \resizebox{.45\textwidth}{!}{
    \begin{tabular}{llll}
    \toprule
    Method & Dataset & Initial Rank& Final Rank \\
    \midrule
    Human & CIRR & 189.98 & 12.34~\textcolor{gray}{$\downarrow_{177.64}$}\\
    Human & FashionIQ & 2427.12 & 447.96~\textcolor{gray}{$\downarrow_{1979.16}$}\\
    Human & FISD & 9.62 & 2.92~\textcolor{gray}{$\downarrow_{6.7}$} \\
    \midrule
    Simulator & CIRR  & 189.98 & 4.12~\textcolor{gray}{$\downarrow_{185.86}$}\\
    Simulator & FashionIQ & 2427.12 & 370.78~\textcolor{gray}{$\downarrow_{2056.34}$}\\
    Simulator & FISD & 9.62 & 4.16~\textcolor{gray}{$\downarrow_{5.46}$} \\
    \bottomrule
    \end{tabular}}}
    \label{tab:user_study}
    \vspace{-10pt}
\end{table}

%% file: tables/ablation_studies.tex
\begin{table}[t]
\scriptsize 
\caption{Ablation studies on CIRR Benchmark based on SPRC model. Bold indicates the best result.}
\label{ablation}
\setlength{\tabcolsep}{1mm}
\centering
\resizebox{\linewidth}{!}{
\begin{tabular}{lccc}
\toprule
Method & Round 1 / Rank$\downarrow$ & Round
3 / Rank$\downarrow$ & Round 5 / Rank$\downarrow$ \\
\midrule
No History & 6.21  & 2.93 & 2.39 \\
Cap. Unchanged & 6.21  & 8.56 & 9.64 \\
Top-10 random & 6.21  & 1.86 & 1.70 \\
Ours & \textbf{6.21}  & \textbf{1.79} & \textbf{1.45} \\

\bottomrule
 \end{tabular}
}
\vspace{-16pt}
\end{table}

%% file: sec/5_conclusion.tex
\section{Conclusion and Future Work}

In this paper, we have illuminated the presence of numerous indeterminate queries within the existing CIR benchmark, which significantly impairs the rigorous evaluation of CIR model performance. To thoroughly assess the capabilities of current CIR models, we introduced FISD, a novel Fully-Informed Semantic-Diverse benchmark tailored for CIR. The experimental results indicate that the current CIR models are inadequate for handling complex composed semantics, particularly in terms of negation and cardinality. Furthermore, we proposed an automated multi-round evaluation framework to assess how CIR models adapt and refine their choices over successive rounds of queries. Extensive experiments and user studies have confirmed the significant improvements achieved by current CIR models in a multi-round setting. We aspire that our proposed benchmark and evaluation framework will provide valuable insights and inspire further research into the CIR problem.

In terms of future work, the current amount of data in our FISD benchmark remains relatively limited. One of the challenges that need to be addressed is how to scale this benchmark. Presently, our multi-round evaluation framework conducts assessments solely on existing CIR models. Therefore, how to develop strong multi-round CIR models is another critical issue that needs to be addressed.

%% file: sec/X_suppl.tex

\onecolumn

{
    \centering
    \Large
    \textbf{A Sanity Check on Composed Image Retrieval}\\
    \vspace{0.5em}Supplementary Material \\
    \vspace{1.0em}
}

\appendix

\section{Details of Generating Different Semantic Subsets}

\noindent \textbf{Addition\&Negation\&Change\&Background.} The generation details for these four semantic subsets are already covered in Section 3.1 of the main paper. In this section, we focus on the specifics of how reference image captions are sampled for each semantic aspect. For the Addition aspect, captions are sampled from the COCO validation set. For Negation, we utilize both the validation and test sets of Flickr30K~\cite{young2014image}. Sampling for the Change aspect is done from the validation set of NoCaps~\cite{agrawal2019nocaps}, while for Background, we draw from the validation set of the CC3M~\cite{sharma2018conceptual} dataset.

Given the image caption for the reference image, we prompt Mixtral-8x7B-Instruct-v0.1~\cite{jiang2024mixtral} to simultaneously generate a relative caption and the caption of the target image with the following prompt:

\begin{figure}[h]
    \centering
    \includegraphics[width=.8\textwidth]{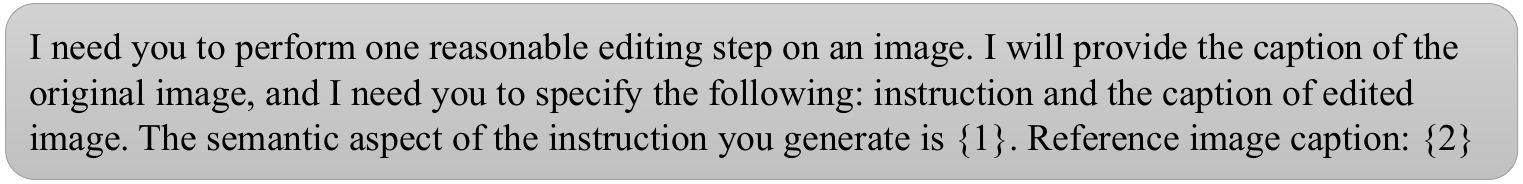}
\end{figure}

where the first placeholder refers to the corresponding semantic aspect, namely addition, negation, change or background, while the second placeholder refers to the reference image caption.

\begin{figure}[h]
    \centering
    \includegraphics[width=.7\textwidth]{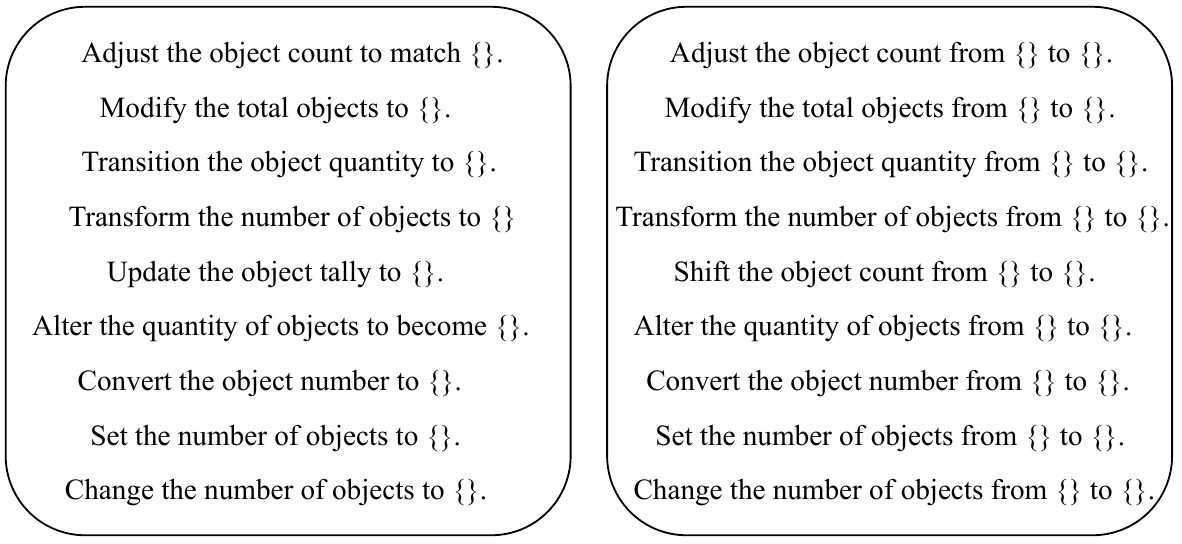}
    \caption{The template for cardinality relative captions.}
    \label{fig:cardinality_template}
\end{figure}

\vspace{2pt}
\noindent \textbf{Cardinality.} We adopt \textit{``a real-life image of \{num\} \{noun\}."} as a template caption, where num is chosen from ten numbers ranging from 1 to 10, and noun is selected from the object category in COCO and ImageNet~\cite{deng2009imagenet}. Then, we employ it to drive the diffusion model to generate corresponding images. After generating the images, we manually adjust the number of objects in the images. Subsequently, we select three images with different numbers of objects belonging to the same category: one serves as a reference image, one as a target image, and the other as a hard negative image. The relative captions are randomly sampled from the templates in Figure~\ref{fig:cardinality_template}. In these templates, the position of the placeholder corresponds to the respective quantity.

\vspace{2pt}
\noindent \textbf{Complex.} We utilize HQ-Edit~\cite{hui2024hq} as the data source for the complex subset. HQ-Edit leverages GPT-4V to provide high-quality reference image captions, relative captions, and target image captions. We select samples with relative captions of more than 25 words as our caption source.

\section{Prompts of Our Multi-round Evaluation Pipeline}

\noindent \textbf{Prompts for MLLM.}
We use MLLM and LLM to simulate user interaction. Specifically, we first prompt MLLM to generate the caption for both candidate and target images. For CIRR, CIRCO, and FISD, we employ the prompt: \textit{Give me a short and precise English description of the image}. For the FashionIQ dataset, which emphasizes clothing, we adopt the following prompt: \textit{Give a short and precise English description of the clothes}.

\vspace{2pt}
\noindent \textbf{Prompts for LLM.} After getting the captions, we prompt LLM to generate relative captions to describe the difference between the candidate and target images. For CIRR and CIRCO, we use the prompt as follows:

\begin{figure}[h]
    \centering
    \includegraphics[width=.8\textwidth]{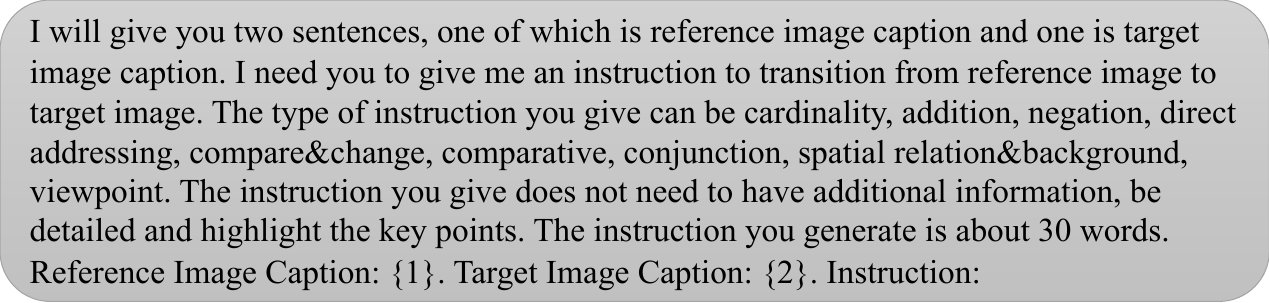}
    \vspace{6pt}
\end{figure}

For FashionIQ, we employ the following prompt that focuses on the semantic aspects potentially contained within the FashionIQ dataset as mentioned in~\cite{liu2021image}.

\begin{figure}[h]
    \centering
    \includegraphics[width=.8\textwidth]{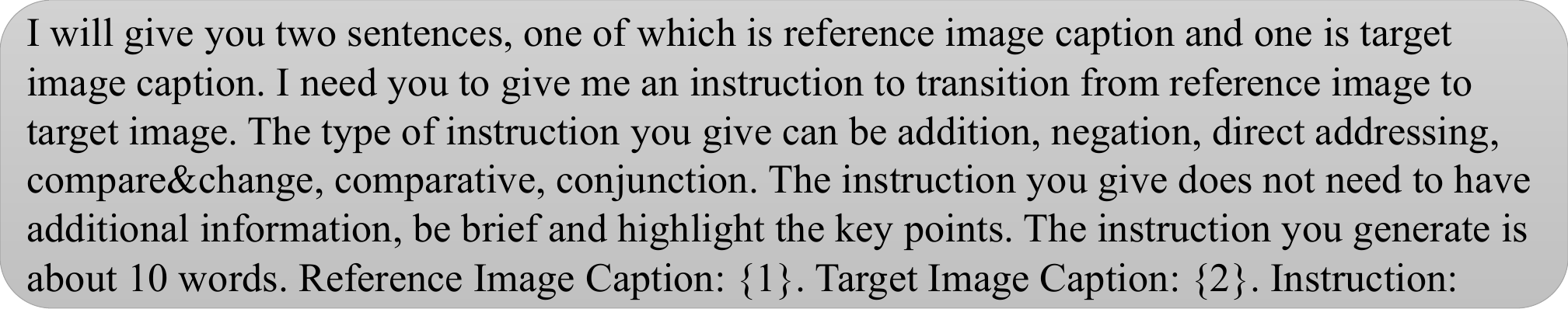}
    \vspace{10pt}
\end{figure}

For FISD, we use the following prompt for addition, negation, cardinality, background, and change semantic aspect, where the first placeholder refers to the specific semantic aspect, the second placeholder denotes the reference image caption, and the third one represents the target image caption.

\begin{figure}[h]
    \centering
    \includegraphics[width=.8\textwidth]{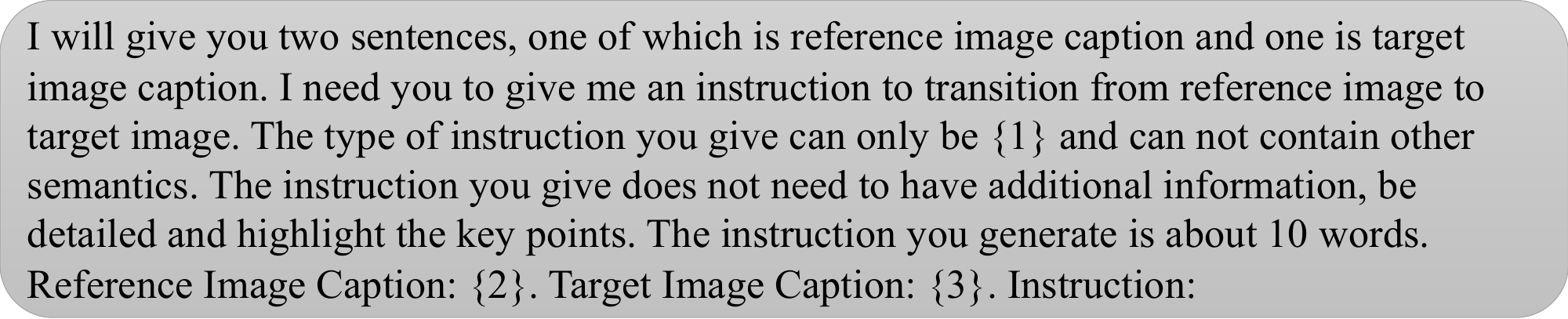}
\end{figure}

Regarding the complex instruction semantic aspect of the FISD benchmark, which may encompass diverse semantic instructions, we do not limit the semantic type. The prompt is as follows: 

\begin{figure}[h]
    \centering
    \includegraphics[width=.8\textwidth]{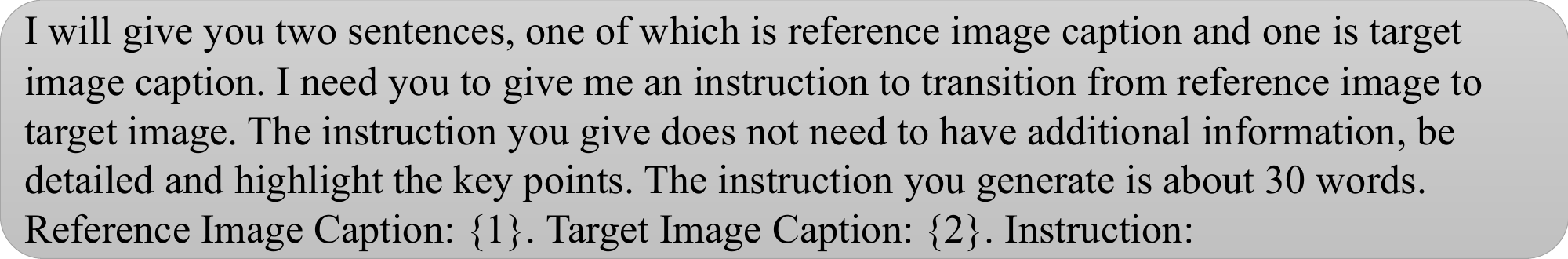}
\end{figure}

\section{Evaluation Details}\label{subsec_supp_implementation_details}

When conducting multi-round evaluations on the FISD, which distinguishes between different semantic aspects, we ensure that the feedback provided by the user simulator contains only the relevant semantic content. For supervised models, when evaluating on the CIRR, CIRCO, and FISD benchmarks, we utilize their official checkpoints, which have been fine-tuned on the CIRR dataset. This is because CIRR, CIRCO, and FISD encompass real-life image scenarios. Conversely, for the evaluation on the FashionIQ dataset, we employ the official checkpoint specifically fine-tuned on the FashionIQ.

\section{Detailed Experimental Results}

\subsection{Additional Validation Experiments for FISD}

\input{supp_tables/cirr_negation}

As shown in Section 4.2 of the main paper, CIR models perform poorly in handling negation semantics on FISD. To determine whether this issue is caused by the gap between synthetic and natural images, we use Llama3-8B to select and manually verify queries with negation semantics from the CIRR validation set. The experimental results are presented in Table~\ref{tab:negation}, revealing that even with natural images, CIR models struggle with negation compared with other semantic aspects. This is consistent with our findings on FISD.

\subsection{Additional Multi-round Evaluation Results}
In this section, we present the multi-round evaluation results of various CIR models using the Recall@K metric. The experimental results are shown in Table~\ref{table:supp_multi_round_eval}.

\input{supp_tables/multi_round_recall_experiments}

\section{Detailed Analysis of User Study}\label{sec_supp_user_study}

\begin{figure}[h]
    \centering
    \includegraphics[width=.6\textwidth]{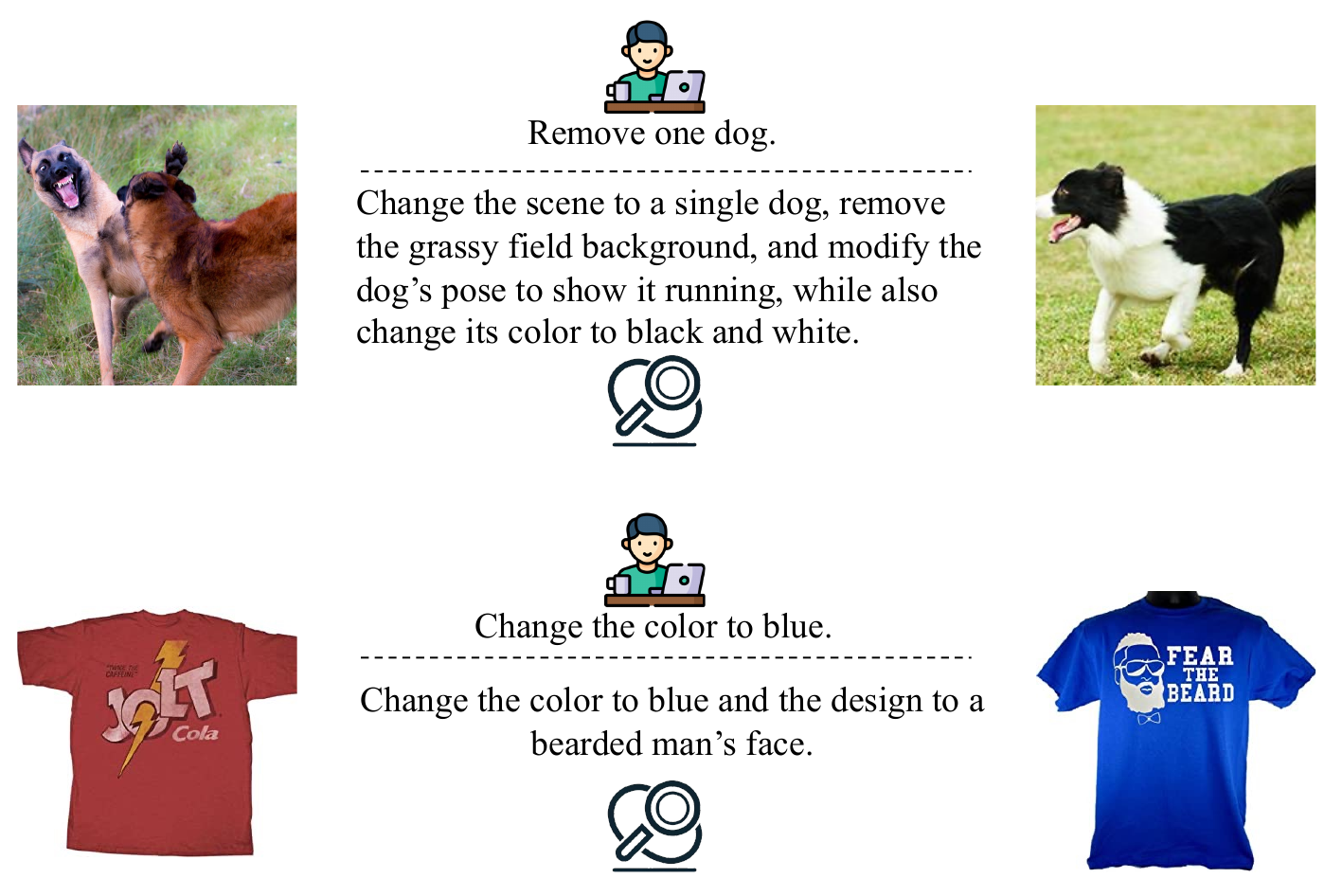}
    \caption{Some examples of feedback from both users and user simulators.}
    \label{fig:supp_user_study}
\end{figure}

\noindent \textbf{Analysis of User Study Experimental Results.} As shown in Section 4.3 of the main paper, in the evaluation of the CIRR and FashionIQ benchmarks, feedback from real users is sometimes less effective compared to that from user simulators. As illustrated in Figure~\ref{fig:supp_user_study}, this disparity may stem from the fact that feedback from real users tends to be concise and lacks details, whereas simulators provide more comprehensive and detailed feedback. 

\vspace{2pt}
\noindent \textbf{Implementation Details of the User Study.} Each time, we present the user with two images: one reference/candidate image and one target image. We ask the user to provide reasonable feedback that describes the differences between the two images. We set the maximum number of interaction rounds to 5, meaning the user provides feedback up to 4 times in one multi-round interaction session. We offer users a wage of \$15 per hour. Note that, our user study is conducted internally within the laboratory, with no potential risks. Additionally, we do not store any feedback provided by the users.

\section{Additional Failure cases in Current Public Benchmark}\label{sec_failure_case_single_round}

\begin{figure}[h]
    \centering
    \includegraphics[width=\textwidth]{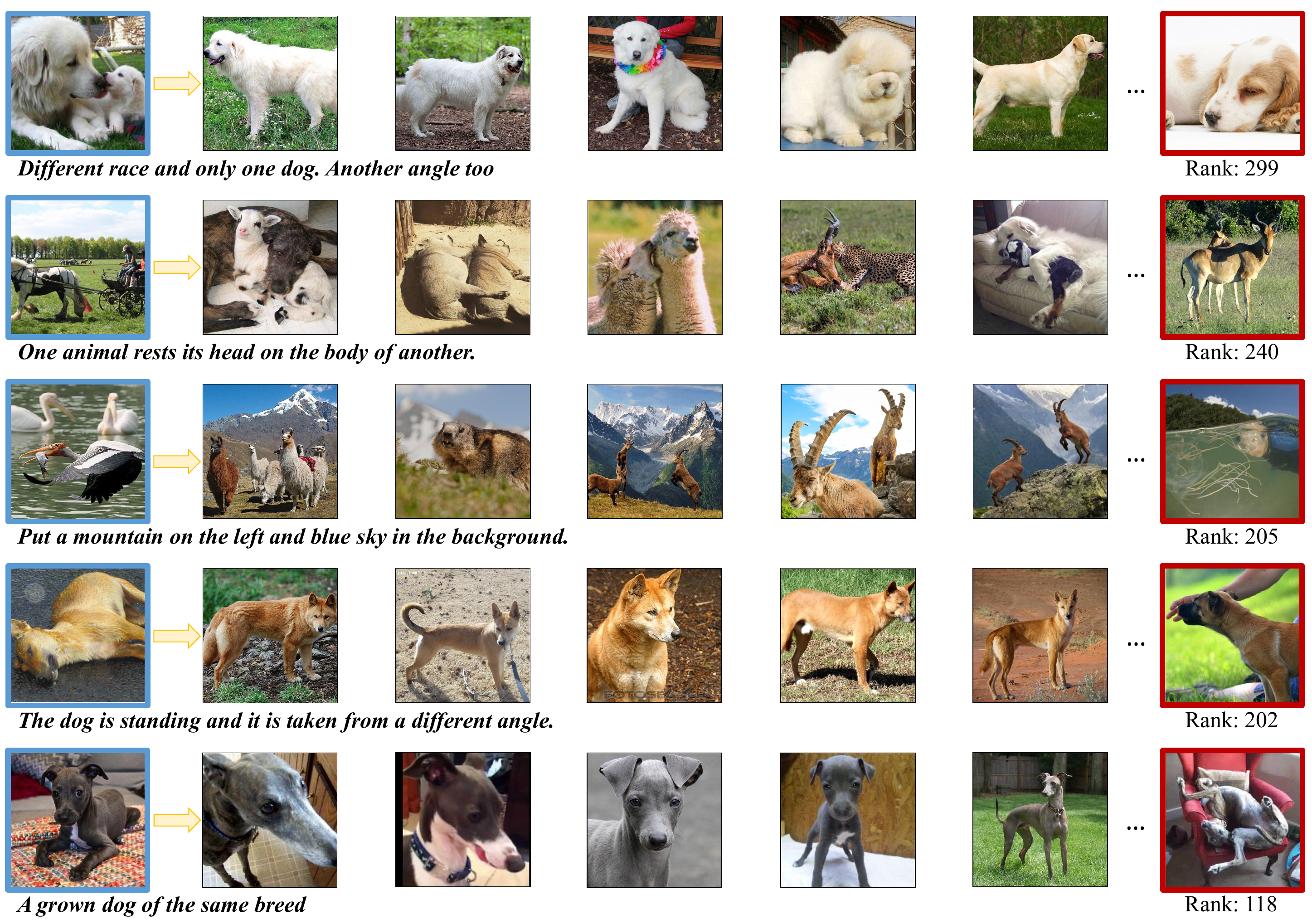}
    \caption{Failure cases of current CIR models on CIRR validation set. The reference image is marked with blue borders and the target image with red borders.}
    \label{fig:supp_benchmark_cirr_failure_cases}
\end{figure}

\begin{figure}[h]
    \centering
    \includegraphics[width=\textwidth]{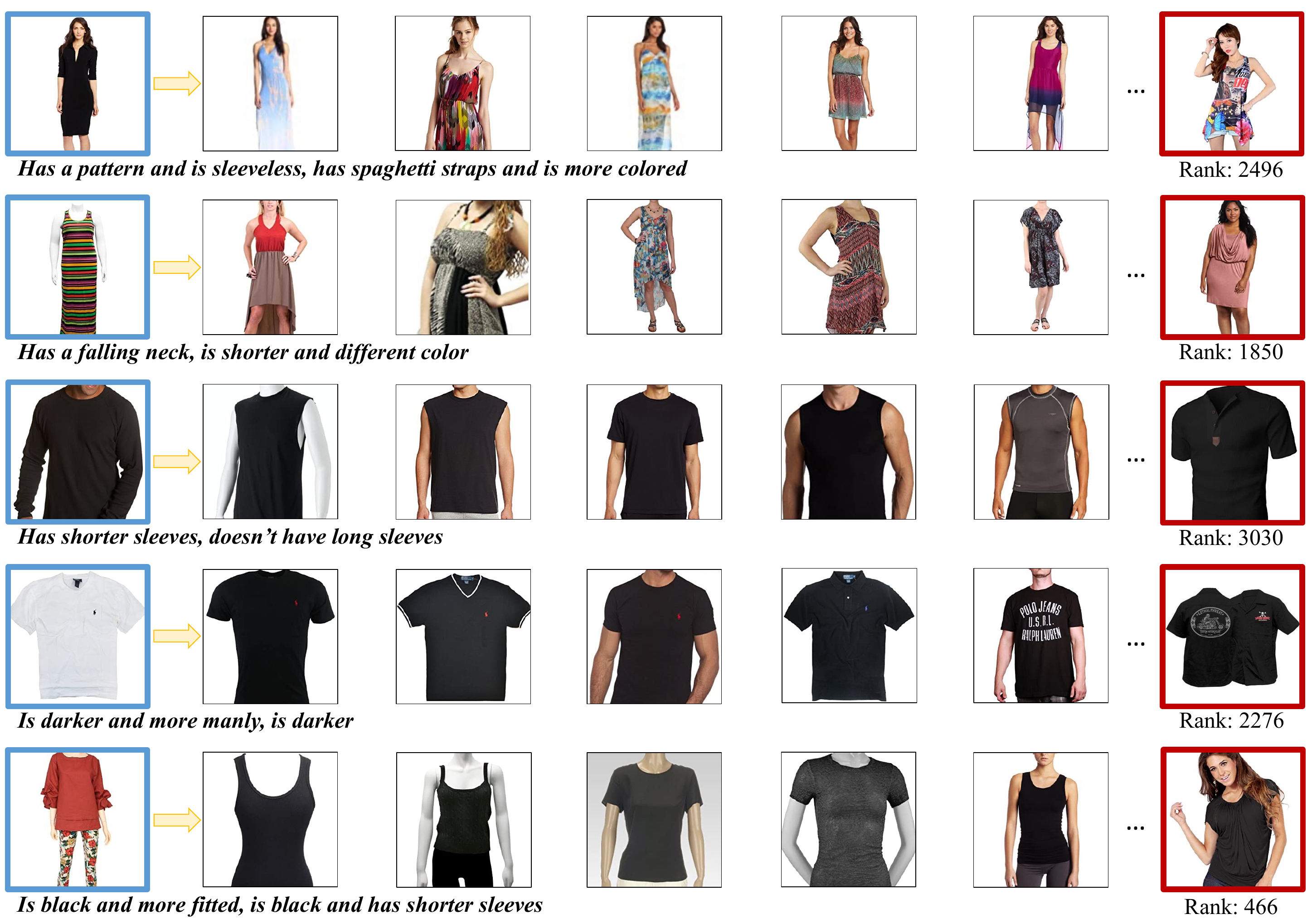}
    \caption{Failure cases of current CIR models on FashionIQ validation set. The reference image is marked with blue borders and the target image with red borders.}
    \label{fig:supp_benchmark_fiq_failure_cases}
\end{figure}

In this section, we show the failure cases of the current single-round CIR models on the CIRR dataset and the FashionIQ dataset. As shown in Figure~\ref{fig:supp_benchmark_cirr_failure_cases} and~\ref{fig:supp_benchmark_fiq_failure_cases}, it can be seen that these failure cases are basically caused by insufficient correlation between the composed query and the target image.

\section{Additional Examples of Our Proposed Benchmark}

\begin{figure}[h]
    \centering
    \includegraphics[width=.85\textwidth]{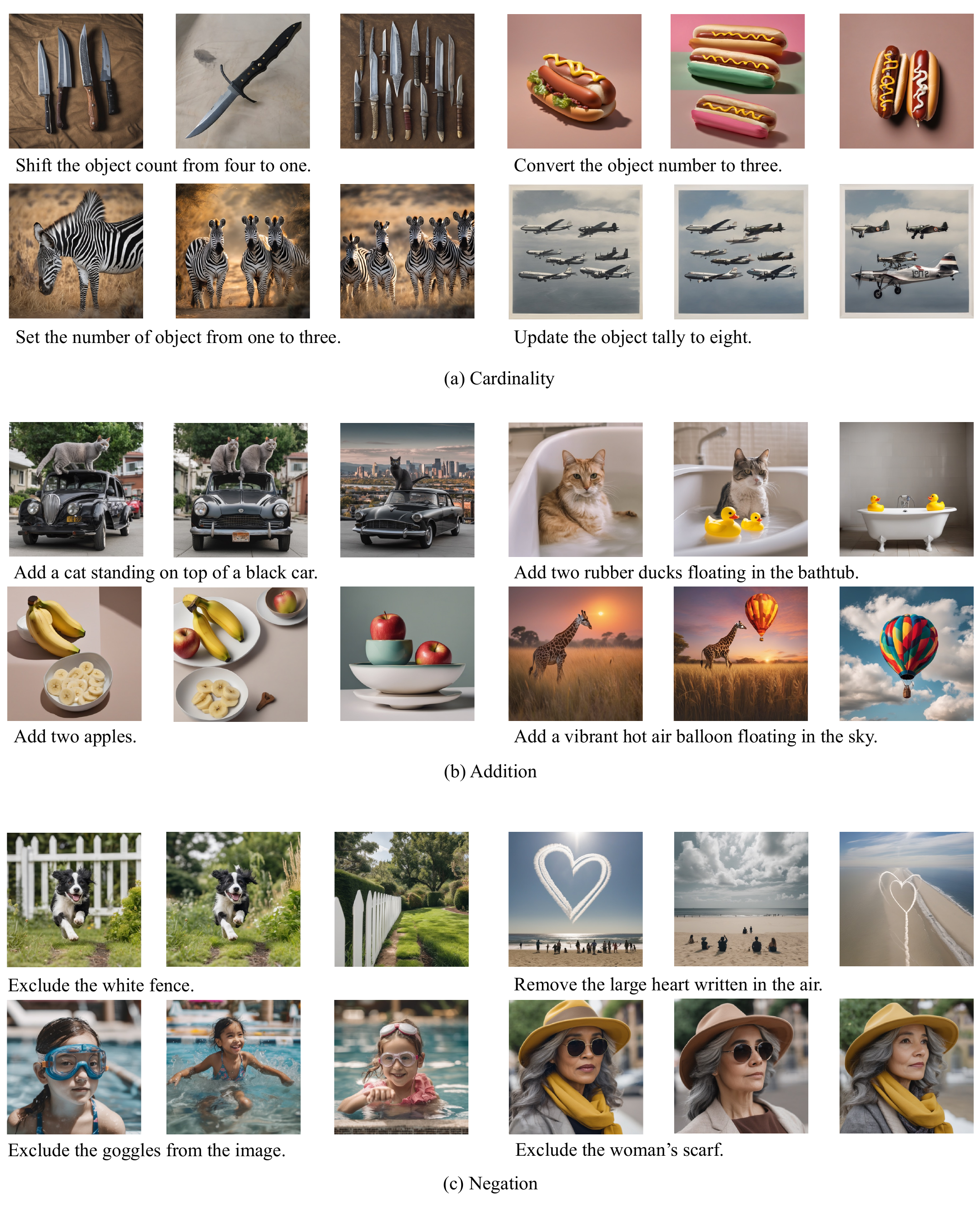}
    \caption{Some examples of our proposed benchmark, the order of the images is a reference image, target image, and hard negative image.}
    \label{fig:benchmark_examples_1}
\end{figure}

\begin{figure}[h]
    \centering
    \includegraphics[width=.85\textwidth]{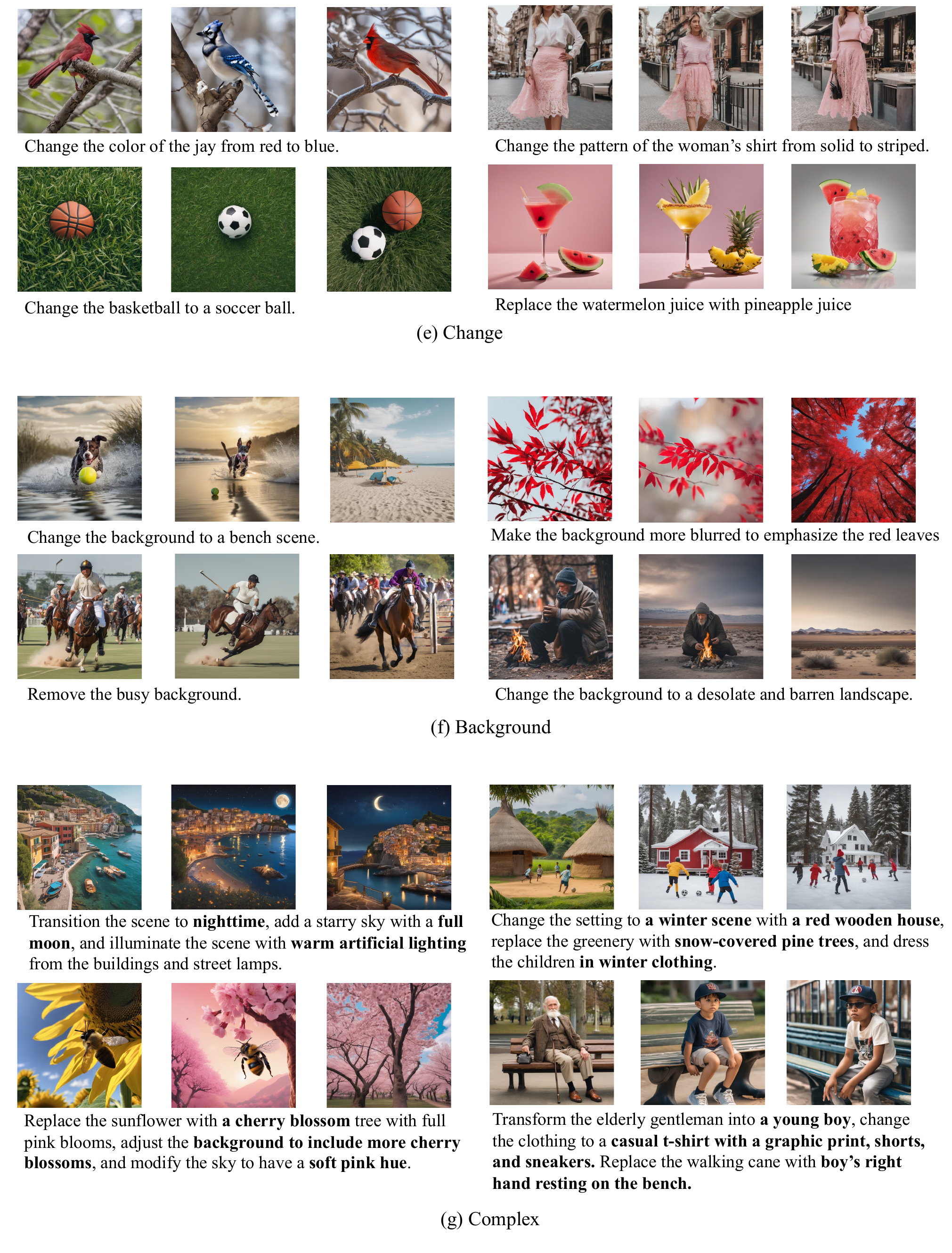}
    \caption{Some examples of our proposed benchmark, the order of the images is a reference image, target image, and hard negative image.}
    \label{fig:benchmark_examples_2}
\end{figure}

In this section, we present additional examples of our proposed FISD benchmark, illustrated in Figure~\ref{fig:benchmark_examples_1} and~\ref{fig:benchmark_examples_2}. These figures display six data types: cardinality, addition, negation, change, background, and complex instruction, respectively. 

\section{Qualitative Analysis}\label{supp_qualitative}

\begin{figure}[h]
    \centering
    \includegraphics[width=0.7\textwidth]{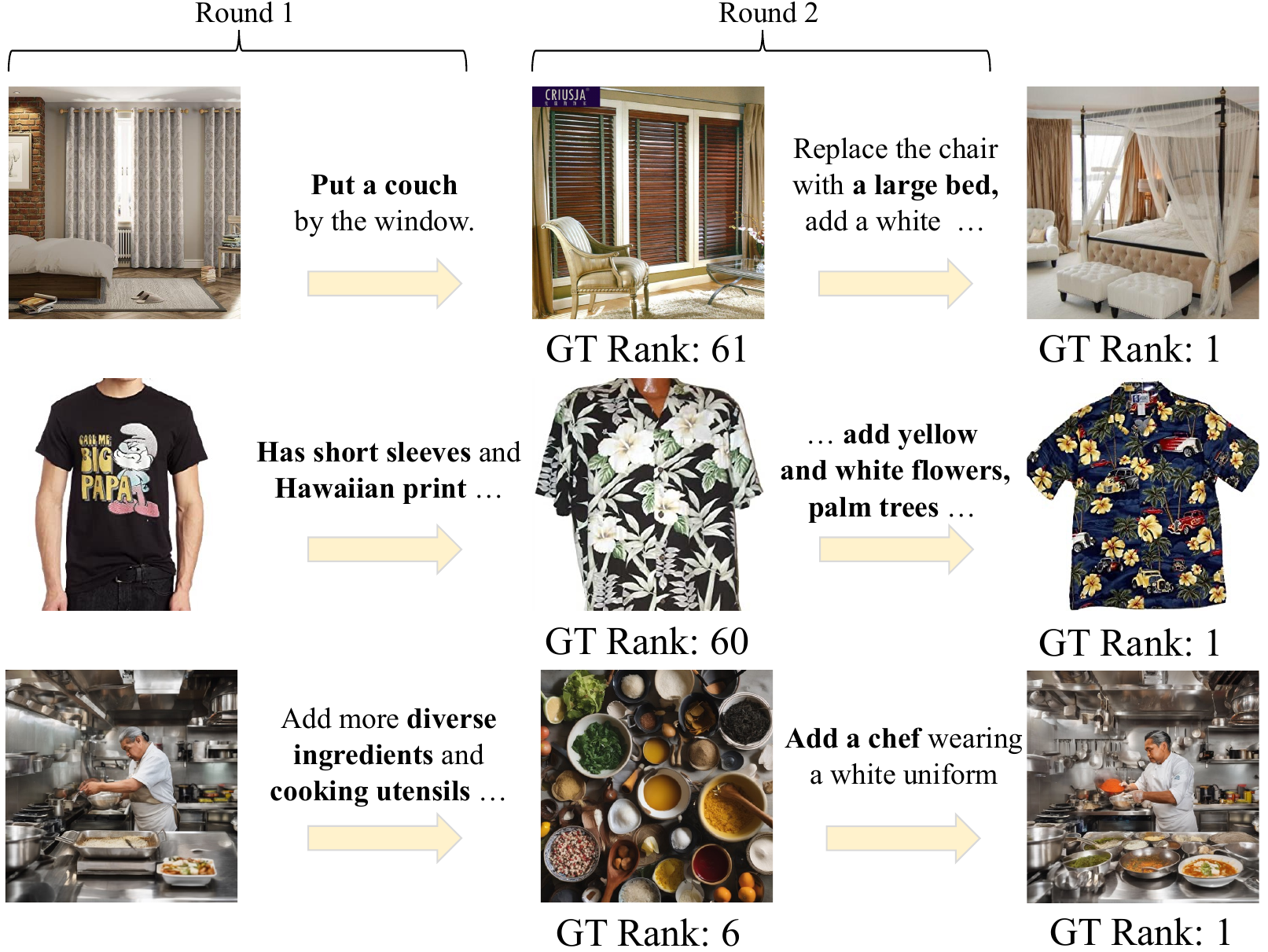}
    \caption{Qualitative results for two rounds.}
    \label{fig:qualitative_two_rounds}
\end{figure}

\begin{figure}[h]
    \centering
    \includegraphics[width=0.9\textwidth]{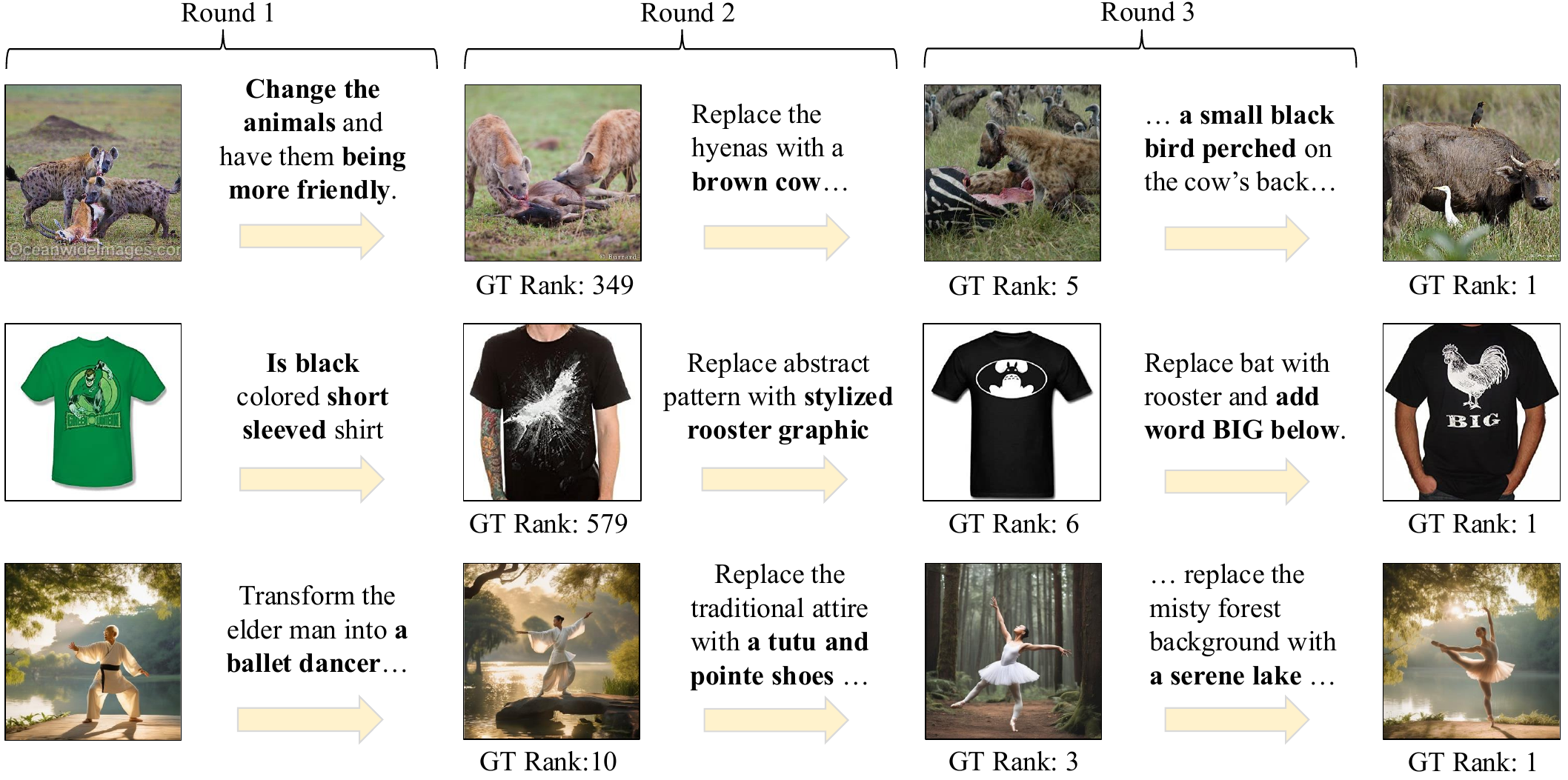}
    \caption{Qualitative results for three rounds.}
    \label{fig:qualitative_three_rounds}
\end{figure}

\begin{figure}[h]
    \centering
    \includegraphics[width=\textwidth]{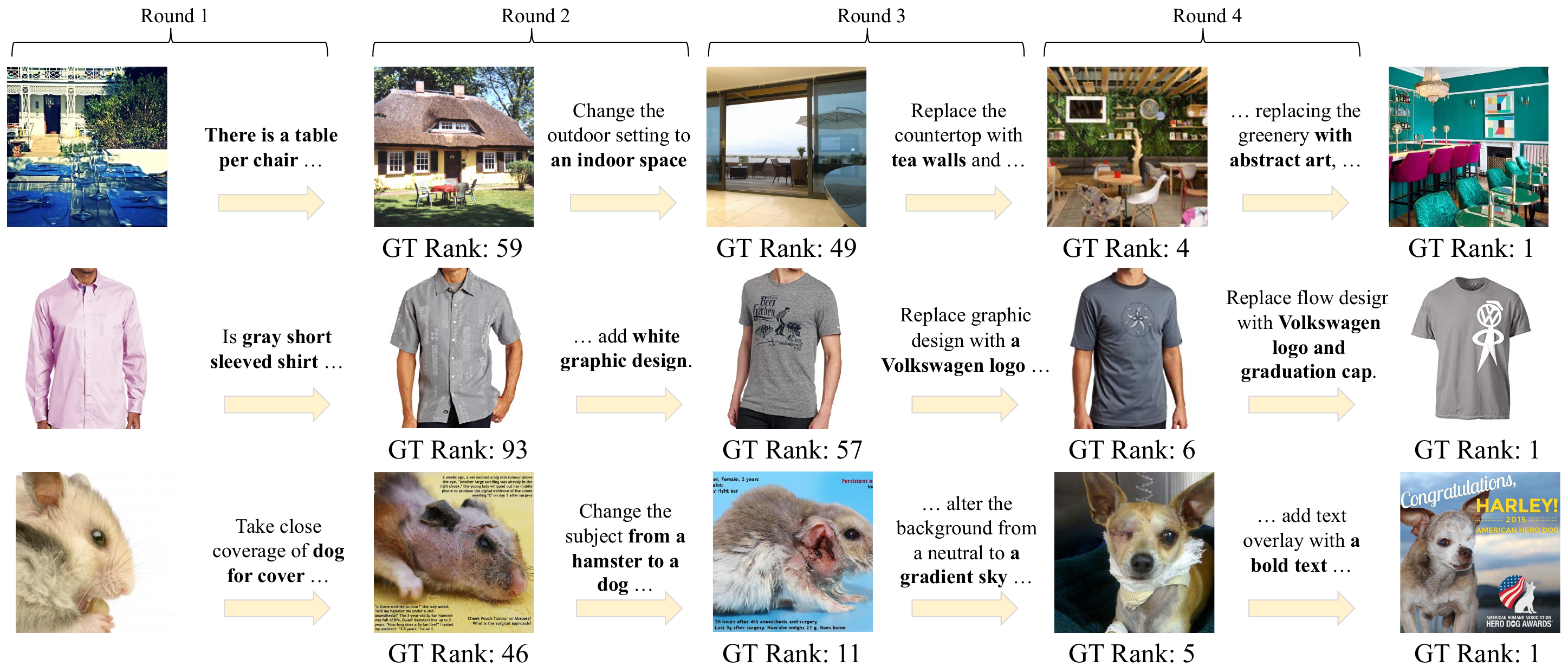}
    \caption{Qualitative results for four rounds.}
    \label{fig:qualitative_four_rounds}
\end{figure}

\begin{figure}[h]
    \centering
    \includegraphics[width=\textwidth]{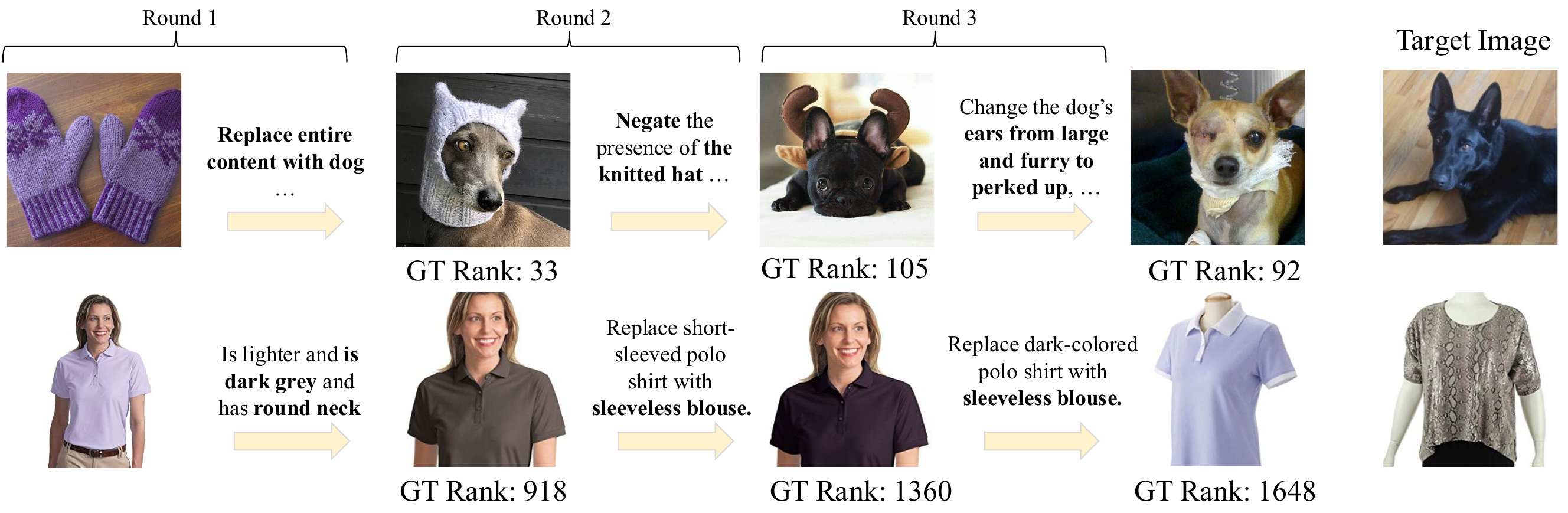}
    \caption{Failure cases in multi-round CIR.}
    \label{fig:supp_multi-round_failure_cases}
\end{figure}

In this section, we show qualitative examples in multi-round CIR. Figure~\ref{fig:qualitative_two_rounds},~\ref{fig:qualitative_three_rounds} and ~\ref{fig:qualitative_four_rounds} illustrate successful results for two to four rounds. These examples demonstrate how our multi-round system incrementally approaches the target image by continuously incorporating user feedback. In Figure~\ref{fig:supp_multi-round_failure_cases}, we showcase some failure cases on the SPRC model under a multi-round setting. These failures primarily occur when receiving either excessively vague feedback or certain semantic feedback types (e.g., negation) that current CIR models can not effectively process. These observations highlight the critical need for both (i) improving the CIR model's overall performance and (ii) developing cleaner, more standardized benchmarks.

\section{Discussion on Inference Cost}
Although the CIR model achieves significant performance gains through multi-round interaction, it inevitably introduces additional overhead. However, in practical applications, there are effective strategies to maximize revenue while minimizing time costs during multi-round explorations. For instance, by dividing the queries into sub-domains, and applying the multi-round exploration only to specific sub-domains that yield better gains, we can reduce the cost of multi-round exploration by shifting from an instance-wise to a sub-domain-wise approach. Moreover, it is also possible to narrow down the candidate pool based on the first-round search, which can also exponentially reduce the time cost in the search space.

%% file: supp_tables/cirr_negation.tex
\begin{table}[h]
\centering
\captionof{table}{Comparison of performance on negation and overall semantics in the CIRR validation set.}
\setlength\tabcolsep{6mm}
    \centering
    \label{tab:supp_elimination}
    \begin{tabular}{c c c}
    \toprule
    \multirow{2}{*}{Method} & Negation & All \\
    \cmidrule(r){2-2} \cmidrule(r){3-3}
    & R@1 & R@1 \\
    \midrule
    Pic2Word & 17.81 & 23.25 \\
    Context-I2W & 20.55  & 26.96 \\
    SPRC & 31.51 & 55.39 \\
    \bottomrule
    \end{tabular}
\label{tab:negation}
\vspace{2pt}
\end{table}

%% file: supp_tables/multi_round_recall_experiments.tex
\begin{table*}[t]
\centering
\scriptsize 
\caption{Multi-round evaluation on various state-of-the-art CIR models across a range of benchmarks.}
\resizebox{\textwidth}{!}{
\setlength{\tabcolsep}{0.26cm}
\begin{tabular}{l c c c c c c c c c}
\toprule
 \multirow{2}{*}[-0.5ex]{Method}&
 \multicolumn{2}{c}{\textbf{FashionIQ-Dress}}& 
 \multicolumn{2}{c}{\textbf{FashionIQ-Shirt}}&
 \multicolumn{2}{c}{\textbf{FashionIQ-Toptee}}&
 \multicolumn{2}{c}{\textbf{CIRR}}&
 \multicolumn{1}{c}{\textbf{FISD}}\\
 \cmidrule(lr){2-3} \cmidrule(lr){4-5} \cmidrule(lr){6-7} \cmidrule(lr){8-9} \cmidrule(lr){10-10}
  & Recall@10 & Recall@50 & Recall@10 & Recall@50 & Recall@10 & Recall@50 & Recall@1 & Recall@5 & Recall@1\\
 \midrule
\multicolumn{10}{c}{\textit{Round 1}} \\
\midrule 
  Pic2Word~\cite{saito2023pic2word} & 20.00 & 40.20 & 26.20 & 43.60 & 27.90 & 47.40 & 23.25 & 51.42 & 21.33 \\
  Context-I2W~\cite{tang2024context} & 23.10 & 45.30 & 29.70 & 48.60 & 30.60 & 52.90 & 26.96 & 56.59 & 27.83\\
  LinCIR~\cite{gu2023language} & 20.92 & 42.44 & 29.10 & 46.81 & 28.81 & 50.18 & 25.09 & 54.41 & 32.42\\
  TransAgg~\cite{liu2023zero} & 30.24 & 51.91 & 34.45 & 53.97 & 38.40 & 59.51 & 38.79 & 69.58 & 34.75\\
  CLIP4CIR~\cite{baldrati2023composed} & 39.46 & 64.55 & 44.41 & 65.26 & 47.48 & 70.98 & 45.37 & 78.47 & 47.17 \\ 
BLIP4CIR+Bi~\cite{liu2024bi} & 42.09 & 67.33 & 41.76 & 64.28 & 46.61 & 70.32 & 42.36 & 75.48 & 45.17  \\
 SPRC~\cite{bai2023sentence} & 49.18 & 72.43 & 55.64 & 73.89 & 59.35 & 78.58 & 55.39 & 84.26 & 50.08  \\ 
 SPN4CIR~\cite{feng2024improving} & 50.57 & 74.12 & 57.70 & 75.27 & 60.84 & 79.96 & 56.47 & 85.29 & 55.83 \\ 
   \midrule
\multicolumn{10}{c}{\textit{Round 3}} \\
\midrule
 Pic2Word~\cite{saito2023pic2word} & 21.81 & 44.42 & 33.12 & 50.93 & 31.72 & 51.50 & 40.28 & 66.01 & 64.58 \\
  Context-I2W~\cite{tang2024context} & 30.84 & 54.69 & 43.62 & 62.66 & 44.97 & 65.83 & 43.00 & 69.31 & 50.08 \\
  LinCIR~\cite{gu2023language} & 24.94 & 48.34 & 45.04 & 62.41 & 39.32 & 60.38 & 42.86 & 70.65 & 50.67\\
  TransAgg~\cite{liu2023zero} & 40.65 & 64.95 & 55.84 & 74.48 & 60.12 & 78.79 & 61.83 & 85.77 & 77.67\\
  CLIP4CIR~\cite{baldrati2023composed} & 46.16 & 69.71 & 59.52 & 79.39 & 61.35 & 80.62 & 67.26 & 93.95 & 74.08\\
BLIP4CIR+Bi~\cite{liu2024bi} & 49.78 & 73.23 & 49.36 & 71.54 & 55.69 & 76.03 & 52.31 & 84.62 & 69.67\\
 SPRC~\cite{bai2023sentence} & 57.91 & 78.78 & 71.05 & 85.92 & 71.49 & 87.35 & 80.63 & 96.22 & 77.08\\
 SPN4CIR~\cite{feng2024improving} & 63.16 & 82.85 & 74.39 & 88.52 & 75.57 & 89.39 & 83.11 & 96.48 & 79.33\\
\midrule
\multicolumn{10}{c}{\textit{Round 5}}\\
\midrule
Pic2Word~\cite{saito2023pic2word} & 22.71 & 43.18 & 34.79 & 51.52 & 32.02 & 51.30 & 57.43 & 68.93 & 83.42 \\
  Context-I2W~\cite{tang2024context} & 31.73 & 54.39 & 46.61 & 65.60 & 47.42 & 68.18 & 56.66 & 71.30 & 68.92 \\
  LinCIR~\cite{gu2023language} & 25.43 & 46.75 & 46.61 & 65.01 & 40.03 & 60.28 & 58.19 & 72.95 & 68.42\\
  TransAgg~\cite{liu2023zero} & 42.69 & 64.90 & 60.99 & 78.56 & 66.29 & 84.14 & 78.02 & 88.23 & 88.33\\
  CLIP4CIR~\cite{baldrati2023composed} & 46.11 & 70.15 & 62.22 & 81.11 & 62.06 & 82.20 & 82.35 & 94.91 & 81.83\\
BLIP4CIR+Bi~\cite{liu2024bi} & 49.98 & 71.99 & 51.37 & 71.59 & 56.45 & 76.29 & 61.97 & 85.22 & 78.83\\
 SPRC~\cite{bai2023sentence} & 57.11 & 76.80 & 73.21 & 86.90 & 73.28 & 87.61 & 88.38 & 97.46 & 83.33\\
 SPN4CIR~\cite{feng2024improving} & 64.20 & 83.74 & 77.53 & 91.41 & 79.30 & 91.59 & 89.26 & 97.54 & 84.92 \\
\bottomrule
\end{tabular}}
\label{table:supp_multi_round_eval}
\end{table*}